\documentclass[conference]{IEEEtran}
\IEEEoverridecommandlockouts
\usepackage{cite}
\usepackage{amsmath,amssymb,amsfonts}
\usepackage{algorithmic}
\usepackage{graphicx}
\usepackage{textcomp}
\usepackage{xcolor}

\usepackage{latexsym}
\usepackage{amsmath}
\usepackage{xcolor}
\usepackage{amsfonts}
\usepackage{mathtools}
\usepackage{ascmac}
\usepackage{algorithm}
\usepackage{algorithmic}
\usepackage{amsthm}  
\usepackage{latexsym}
\usepackage{arydshln}
\usepackage{bm}
\usepackage[subrefformat=parens]{subcaption}
\usepackage{color}
\usepackage{colortbl}
\usepackage{multirow}
\usepackage{url}
\usepackage{etoolbox} 

\def\BibTeX{{\rm B\kern-.05em{\sc i\kern-.025em b}\kern-.08em
    T\kern-.1667em\lower.7ex\hbox{E}\kern-.125emX}}
\begin{document}

\title{Bounding the Worst-class Error: A Boosting Approach\\
\thanks{This work was supported in part by JSPS KAKENHI JP23KJ1723, JP24K03002, 24K22308, 22H05180, and JST ACT-X JPMJAX23CR.}
}

\makeatletter
\patchcmd{\@IEEEauthorblockA}{\relax}{\vspace{-0.8ex}}{}{}
\makeatother

\author{
Yuya Saito\IEEEauthorrefmark{1},
Shinnosuke Matsuo\IEEEauthorrefmark{1},
Seiichi Uchida\IEEEauthorrefmark{1},
Daiki Suehiro\IEEEauthorrefmark{1}\IEEEauthorrefmark{2} \\
\IEEEauthorblockA{\IEEEauthorrefmark{1}Department of Information Science and Technology, Kyushu University, Japan, \IEEEauthorrefmark{2}RIKEN AIP, Tokyo, Japan} 
\texttt{\{yuya.saito@human., shinnosuke.matsuo@human., uchida@, suehiro@ \}ait.kyushu-u.ac.jp}
}

\def\BibTeX{{\rm B\kern-.05em{\sc i\kern-.025em b}\kern-.08em
    T\kern-.1667em\lower.7ex\hbox{E}\kern-.125emX}}

\maketitle
\newcommand{\figcaption}[1]{\def\@captype{figure}\caption{#1}}
\newcommand{\tblcaption}[1]{\def\@captype{table}\caption{#1}}
\makeatother


\newcommand{\C}[1]{{#1}}
\newcommand{\bb}[1]{\mathbb{#1}}

\newcommand{\red}[1]{\textcolor{red}{#1}}
\newcommand{\green}[1]{\textcolor{green}{#1}}
\newcommand{\blue}[1]{\textcolor{blue}{#1}}


\renewcommand{\algorithmicrequire}{\textbf{Inputs:}}
\renewcommand{\algorithmicensure}{\textbf{Output:}}

\newtheorem{defi}{Definition}
\newtheorem{theo}{Theorem}
\newtheorem{prop}[theo]{Proposition}
\newtheorem{coro}[theo]{Corollary}
\newtheorem{rem}[theo]{Remark}
\newtheorem{lemm}{Lemma}
\newcommand{\Rdm}{\mathfrak{R}}

\begin{abstract}
This paper tackles the problem of the worst-class error rate, instead of the standard error rate averaged over all classes. For example, a three-class classification task with class-wise error rates of 10\%, 10\%, and 40\% has a worst-class error rate of 40\%, whereas the average is 20\% under the class-balanced condition.
The worst-class error is important in many applications. For example, in a medical image classification task, it would not be acceptable for the malignant tumor class to have a 40\% error rate, while the benign and healthy classes have a 10\% error rates. To avoid overfitting in worst-class error minimization using Deep Neural Networks (DNNs), we design a problem formulation for bounding the worst-class error instead of achieving zero worst-class error. Moreover, to correctly bound the worst-class error, we propose a boosting approach which ensembles DNNs.
We give training and generalization worst-class-error bound.
Experimental results show that the algorithm lowers worst-class test error rates while avoiding overfitting to the training set. This code is available at \url{https://github.com/saito-yuya/Bounding-the-Worst-class-Error-A-Boosting-Approach}. 

\end{abstract}


\section{Introduction}
\label{sec:intro}

\begin{figure}[t]
\centering
    \centering
    \includegraphics[keepaspectratio, width=1.0\linewidth]
    {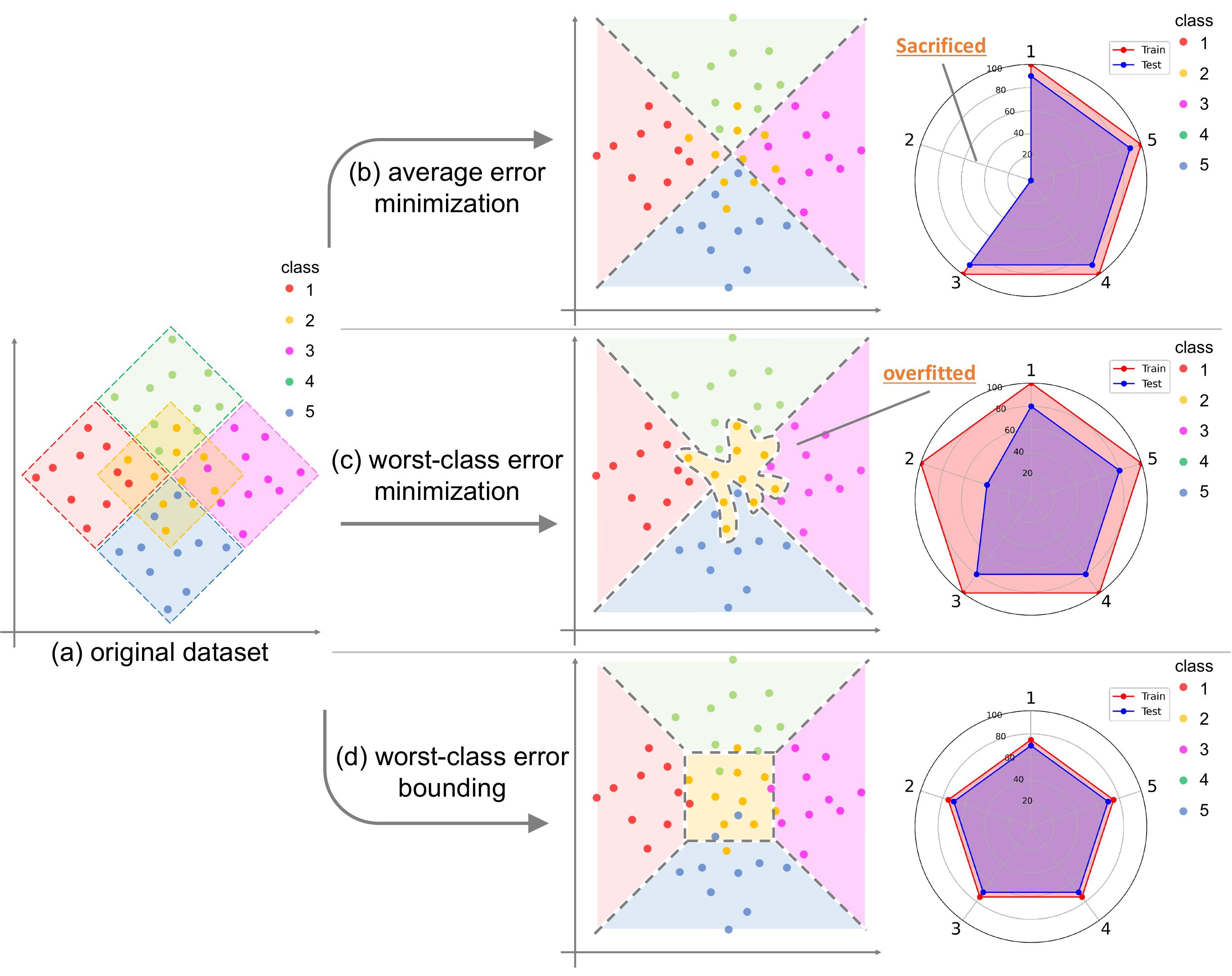}

    \caption{A toy dataset that shows how our worst-class error bounding provides a reasonable class boundary.}
    \label{fig:balance_toydata}
\end{figure}

Minimizing the {\em average} training error over all classes can lead to large differences in class-wise errors. Fig.~\ref{fig:balance_toydata}{(a)} shows a toy example of a five-class problem.
If we aim to achieve a small test average error, Fig.~\ref{fig:balance_toydata}{(b)} can be a possible good boundary because class 2 (yellow) overlaps with the other classes.
However, class 2 has a very high error rate, whereas the other classes have almost zero;  that is, class 2 is sacrificed for the average. This example indicates that the average error is not suitable for problems such as medical image classification, where no class should be sacrificed.
For example, in a medical image classification task, it would not be acceptable for the malignant tumor class to have a 40\% error rate while the benign and healthy classes have 10\% error rates.
\par
To avoid sacrificed classes, minimizing the training {\em worst-class} error is more appropriate rather than the average error. Let us consider a $K$-class classification with a training sample of instance-label pairs $S = ((x_1, y_1), \ldots, (x_n, y_n)) \in (X\times Y)^n$ and denote class-wise instances by $\{S_1, \ldots, S_K\}$, where $S_j=\{x \mid (x,y) \in S, y=j \}$. the training worst-class error of a hypothesis $h$ for $S$ is defined as:
\begin{equation}
\label{eq:worst-class}
{\hat{U}}_{{S}}(h) = \max_{k \in [K]}{\hat{R}}_{S_k}(h),
\end{equation}
where $[K]=\{1, \ldots, K\}$ and $\hat{R}_{S_k}(h)$ is the training error rate of class $k$ by $h$, i.e., ${\hat{R}}_{S_k}(h) = \frac{1}{n_k} \sum_{x \in S_k} I(h(x) \neq k)$, where $I$ is the indicator function and $n_k=|S_k|$.
Note that Eq.~\eqref{eq:worst-class} is a general formulation, and thus, we can consider any hypothesis set $H$ and find $h \in H$ with some appropriate surrogate loss function of classification error. Also, note that high worst-class errors occur not only in class-imbalanced data but also in class-balanced data like Fig.~\ref{fig:balance_toydata}.
Despite its importance in various applications, the learning problem for worst-class error has not been discussed yet, to the author's best knowledge. 
\par
Suppose that we find $h \in H$ that minimizes the training worst-class error ${\hat{U}}_{{S}}(h)$ with a rich hypothesis set $H$, such as deep neural networks (DNNs). As shown in Fig.~\ref{fig:balance_toydata}{(c)}, DNNs will easily achieve ${\hat{U}}_{{S}}(h)=0$ (i.e., the training worst-class error becomes zero) due to the high representation ability. However, this hypothesis $h$ is just overfitted to $S$ and will not perform well on a test set.
\par
This paper proposes a more practical formulation for {\em bounding} the worst-class error. Specifically, instead of minimizing ${\hat{U}}_{{S}}(h)$ as much as possible, we aim to find $h$ to satisfy an upper bound for the training worst-class error,  
\begin{align}
\label{eq:bounding_trerr}
{\hat{U}}_{{S}}(h) < 1- \theta, 
\end{align}
where $\theta \in [0,1)$ is a predefined hyperparameter that specifies the upper bound of the training worst-class error. Overfitting can be avoided by allowing up to $1-\theta$ training worst-class errors, as shown in Fig.~\ref{fig:balance_toydata}{(d)}.
Note that, depending on the data and hypothesis set, no $h$ can satisfy Eq.~\eqref{eq:bounding_trerr}. Practically, we need to determine an appropriate $\theta$ as detailed in Section~\ref{sec:exp}.
\par
This paper also proposes a boosting algorithm for finding an ensemble hypothesis that satisfies Eq.~\eqref{eq:bounding_trerr}. The following two points characterize our boosting. First, it minimizes a zero-one loss  $\ell_k^\theta  \in \{0,1\}$, which judges whether the class-wise error of $S_k$ is less than $1-\theta$. Second, our algorithm assigns the weights to individual {\em classes}, whereas the standard boosting algorithms assign the weights to individual {\em instances}.
\par
We prove that our boosting algorithm theoretically guarantees that the training worst-class error is bounded by $\theta$ after a sufficient number of rounds $T$. A remarkable result is that the worst-class generalization error depends on the sample size of the hardest class to distinguish or the smallest class. It is important that, for class-imbalanced data, the difficulty of the problem is often defined by the class-imbalance ratio (i.e., the sample ratio between the largest class and smallest class)~\cite{buda2018systematic,CB,IB}. In contrast, our theorem says that the worst-class error can be bounded when we have a sufficiently large number of training instances over any class, even when the sample is highly class-imbalanced.
\par
The main contributions are summarized as follows:
\begin{itemize}
    \item We propose a novel learning problem that focuses on the worst-class error.
    \item We propose the problem reformulation for bounding the worst-class error but not just minimizing it, and effectively solve it by a boosting algorithm.
    \item We derive theoretical guarantees of convergence for the training worst-class error and the bound of generalization worst-class error of the proposed boosting algorithm. Remarkably, the provided generalization bound gives the new notion that the difficulty of the class-imbalanced problem is characterized by the sample size of the hardest/smallest class but not the imbalance ratio. 
    \item Experimental results show that our algorithm outperforms state-of-the-art methods for balanced and imbalanced scenarios while achieving a small worst-class error.
\end{itemize}

\section{Related work}
\subsection{Class-wise error and its bias}
Class-wise error is an important metric in various fields. For example, if we find a class with a large class-wise error in a medical image classification task, we readily understand that this classifier is risky in its practical use (e.g.,~\cite{giotis2015med,hassan2020developing}).
\par
Class-wise error is a popular evaluation criterion for class-imbalanced tasks.
~\cite{IB} focused on the performance of class-wise error in their experiment and showed the performance improvement of the minority class.
There are various techniques for dealing with imbalanced tasks.
In particular, the focal loss~\cite{focal_loss} can increase the weight for hard instances and decrease the weight for easy instances, whereas the class-balanced loss~\cite{CB} weights the loss of each class according to the effective number of instances, not just the number of class-wise instances.
The influence-balanced loss~\cite{IB} can decrease the effect of the instances that induce overfitting.
In addition, numerous methods have been proposed recently, such as~\cite{LA, CDT}.
However, as mentioned in Section~\ref{sec:intro},
the bias of the class-wise error is a potential problem not only for class-imbalanced tasks but also for class-balanced tasks. We are not sure whether such bias can also be avoided in class-balanced tasks.
\par
Group-wise error has also been extensively studied due to the growing attention to fairness.
\cite{VSloss} considers group-sensitive classification tasks.
It showed that the proposed method, called VS-loss, is also
effective for class-imbalanced data.
The generalization analysis on the balanced error was provided for specific models and data.
\cite{chaudhuri2023does}
proposed generalization analysis on worst-group/class error for linear classifiers. However, as aforementioned, overfitting is a crucial problem when using DNNs. 
Our survey of the literature has not yielded sufficient discussion on the worst-class error or its theoretical and practical performance for DNNs.

\subsection{Boosting}
\label{subsec:boost}
Boosting algorithm~\cite{adaboost} learns an ensemble hypothesis (or strong learner) by using a weak learning algorithm.
Boosting is an iterative procedure. At each round $t \in [T]$, the weak learning algorithm returns a hypothesis $h_t$ and updates the weights (or distribution) of the training instances such that the performance of $h_t$ becomes low. After $T$ rounds, the algorithm returns an ensemble hypothesis of $h_1, \ldots, h_T$.
Boosting is both theoretically motivated and practically effective, and it is widely used in various learning tasks.
\par
There are some boosting algorithms for dealing with imbalance-tasks~\cite{galar2011review,tanha2020boosting}.
However, to the best of our knowledge, there is no theoretical or practical boosting algorithm for bounding the worst-class error.
Therefore, in this study, we modify the definition of weak learnability specialized for 
theoretically bounding the worst-class error (see Section~\ref{sec:weak_learn}) and propose a practical way to use a DNN as a weak learner based on the definition of weak learnability (see Section~\ref{sec:exp}). 
\par
Recently, it has been shown that a boosting algorithm for binary classification tasks can be reduced to Online Convex Optimization (OCO)~\cite{OCO}.
The reduction technique is that a standard boosting algorithm can be considered to be a repeated game between a weak learning algorithm $\mathcal{A}$ and an instance-weighting algorithm $\mathcal{B}$ for updating the weights of instances.
The repeated game is given as follows:
in each round $t \in [T]$, $\mathcal{B}$ predicts a vector $\boldsymbol{v}_t \in \Delta_n$ (weights of the classification loss), the weak learning algorithm finds a hypothesis $h_t$ corresponding to the weighted instances and gives a feedback vector $\boldsymbol{r}_t \in \{0,1\}$ (classification performance of $h_t$), where $n$ is the number of instances and $\Delta_n$ is a $n$-dimensional probability simplex. 
$\mathcal{B}$ is designed to achieve a small \emph{regret}: 

\begin{align}
\label{align:regret}
\mathrm{Regret}(\mathcal{B})= \sum_{t=1}^T \boldsymbol{v}_t \cdot \boldsymbol{r}_t - \min_{\boldsymbol{v} \in \Delta_n}\sum_{t=1}^T \boldsymbol{v} \cdot \boldsymbol{r}_t 
\end{align}
In the original OCO boosting, we can use zero-one classification loss, and it achieves zero classification training error with the theoretical guarantee. 
In this study, on the basis of the OCO boosting framework, we design a novel and simple boosting algorithm specialized for bounding the worst-class error. Consequently, we can simply minimize the proposed zero-one loss for bounding the worst-class error. (Note that a well-known AdaBoost~\cite{adaboost} minimizes not the zero-one loss but the logistic loss.)

\section{Preliminaries}
Let $X$ be an instance space and $Y = [K]$ be a label space, and $H \subseteq \{h: X \to Y\}$ be a hypothesis set. The learner has a sample $S$
drawn i.i.d. according to an unknown distribution $D$. In multi-class learning problems,
the goal of the learner is to find $h \in H$ so that the generalization average error
${R_D}(h) = \bb{E}_{(x,y) \sim {D}} [I(h(x) \neq y)]$
is small. The training average error is defined as $\hat{R}_S(h) = \frac{1}{n} \sum_{(x,y) \in S} I(h(x) \neq y)$.
We denote an ensemble hypothesis by $\bar{h}$. Specifically, we consider the majority vote of 
$T$ hypotheses over $H$. The hypothesis set is defined as $\bar{H}=\{\bar{h}: x \mapsto \arg\max_{k \in [K]}\sum_{t=1}^T I({h}_t(x)=k) \mid h_t \in H \}$.

\section{Formulations for bounding worst-class error}
We consider the worst-class error instead of the average error. The generalization class-wise error of $h \in H$ is defined as ${R}_{{D}_k}(h) = \bb{E}_{x \sim {D}_k} [I(h(x) \neq k)]$, 
where ${D}_k = {D}(x \mid y=k)$. The generalization worst-class error of $h$ is defined as:
\begin{align}
\label{align:gen_err}
{U}_{{D}}(h) = \max_{k \in [K]} {R}_{{D}_k}(h).
\end{align}
Similarly, for the training sample $S$, the training worst-class error is defined as~\eqref{eq:worst-class}.
 A naive solution to find $h$ is a direct minimization of ${\hat{U}}_{{S}}(h)$ with a surrogate loss. However, it may induce overfitting, as we discussed in Section~\ref{sec:intro}. Especially when we employ a rich hypothesis set such as a DNN, the training worst-class error can be zero, as shown in Fig.~\ref{fig:balance_toydata}~(c).
\par  
In this paper, we find $h \in H$ which satisfies Eq.~\eqref{eq:bounding_trerr}, i.e.,
${\hat{U}}_{{S}}(h) < 1- \theta,$ 
for upper-bounding the worst-class error by $1-\theta$, where $\theta\in [0, 1)$ is a hyperparameter. In other words, we find $h$ tolerant to $1-\theta$ class-wise errors. With this upper-bounding, we avoid overfitting and achieve smaller ${U}_{{D}}(h)$ like Fig.~\ref{fig:balance_toydata}~(d).
\par  
A possible strategy is to consider some surrogate loss $\mathcal{L}: X \times Y \times H \to \mathbb{R}$ (e.g., hinge-loss, cross-entropy loss) of zero-one classification error (i.e., $I(h(x)\neq y)$) and attempt to bound the class-wise surrogate loss by $\theta' \in \mathbb{R}^+$, i.e., a possible goal is to find $h$ such that
 $\hat{U}_S^{\mathcal{L}}(h)=\frac{1}{n_k}\max_{k \in [K]}\sum_{x \in S_k}\mathcal{L}(x,k,h) < \theta'$. However, it is difficult to derive $\theta'$ from a user-defined target $\theta$. Moreover, there may be the difference of ``worst class'' between the zero-one classification error and its surrogate loss because the surrogate loss possibly takes large values with hard misclassification and takes positive values even with correct classification.

\section{Bounding worst-class error by boosting} 
First, we reformulate the problem of finding $h$ satisfying Eq.~\eqref{eq:bounding_trerr} as the minimization of a class-wise zero-one loss $\ell_{k}^\theta$ for any $k \in [K]$.
Second, motivated by the application of boosting for minimizing a standard zero-one classification loss~\cite{zhai2013direct}, we provide a boosting algorithm for minimizing $\ell_{k}^\theta$.
\subsection{Reformulation of worst-class error bounding and boosting}
The problem of Eq.~\eqref{eq:bounding_trerr} is reformulated as an equivalent problem of finding $h$ that satisfies

\begin{align}
\label{align:objective}
{\ell}_{k}^\theta({h})=0,\ \forall k \in [K], 
\end{align}
where ${\ell}_{k}^\theta({h})$ is a zero-one loss:

\begin{align}
\label{align:zeroone_train_err}
~ {\ell}_{k}^\theta({h})=&I({\hat{R}}_{S_k}({h}) \geq 1-\theta) \\ \nonumber
=&I\left(\frac{1}{n_k} \sum_{x \in S_k} I({h}(x) \neq k) \geq 1-\theta \right).
\end{align}
Since the zero-one loss is not convex and non-smooth, it cannot be solved by standard optimization approaches. We, therefore, utilize a boosting framework of~\cite{OCO} to minimize the zero-one loss directly like~\cite{zhai2013direct}. More precisely, our goal is to find $\bar{h} \in \bar{H}$ which minimizes ${\ell}_{k}^\theta({\bar{h}}),\ \forall k \in [K]$. 
Note that our boosting can return $\bar{h}$ satisfying ${\ell}_{k}^\theta({\bar{h}})=0,\ \forall k \in [K]$ through the minimization, as proved in Section~\ref{subsec:training_theo}.
\subsection{Weak learnability}
\label{sec:weak_learn}
In general, the boosting algorithm is guaranteed to obtain a strong learner under weak learnability, which is the assumption of weak learning algorithms. The weak learnability specialized to our boosting is defined as follows.
\begin{defi}[weak learnability]
Let $\mathcal{A}$ be an algorithm that inputs $S$ and returns $h \in H$.
$H$ is said to be $(\gamma, \delta, \theta)$-weak learnable if there exists an algorithm $\mathcal{A}$, $\gamma>0$ and a polynomial function $poly(\cdot, \cdot)$ that satisfies: for any $\delta>0$ and for a distribution $D$ with any class prior probabilities $\{D_1, \ldots, D_K\}$, the following holds for any sample size $n \geq poly(\delta, \gamma)$:
\begin{align}
\label{def:weak-learnability}
    \bb{P}_{S\sim D^n} \left[\frac{1}{K}\sum_{k=1}^K \ell_{k}^\theta({\mathcal{A}}(S)) \leq \frac{1}{2} - \gamma \right] \geq 1- \delta.
\end{align}
\end{defi}
\noindent This simply means that for any class prior probabilities, the weak learning algorithm $\mathcal{A}$ returns $h$ such that
$\frac{1}{K}\sum_{k=1}^K {\ell}_{k}^\theta(h)$ is less than $1/2$.

\begin{algorithm}[t]
    \caption{Our boosting algorithm for bounding the worst-class error}
    \label{alg1}
    \begin{algorithmic}[1]  
    \REQUIRE $ {H}, \gamma, \delta, 
    \theta \in [0,1), {S}, T, \eta>0$  
    \ENSURE $\bar{h}$
    \STATE{$\text{Set initial weights:} \: \boldsymbol{w}_1 = \frac{1}{K} \boldsymbol{1}$}
    \FOR {$t = 1,2,\ldots,T$}
    \STATE {Find hypothesis by the $(\gamma,\delta/T,\theta)$-weak learning algorithm:}  

    \begin{align}
    \label{align:weighted_err}
        h_t = \arg\min_{h \in H}\sum_{k=1}^K w_{t,k}\hat{R}_{S_k}(h)
\end{align}
    \STATE Define the feedback vector:
    \begin{align}
    \label{align:feedback}
    r_{t,k}(h_t) = 1- \ell_k^\theta(h_t)
    \end{align}
    \STATE Update each weight:
    \begin{align}
    \label{align:hedge}
    {w}_{t+1,k}= \frac{w_{t,k}e^{-\eta r_{t,k}(h_t)}}{\sum_{j=1}^{K}w_{t,j}e^{-\eta r_{t,j}(h_t)}}
    \end{align}
    \ENDFOR
    \end{algorithmic}
\end{algorithm}
\par

\subsection{Boosting algorithm\label{sec:algorithm} for bounding worst-class error} 
\noindent\textbf{Outline}: 
Algorithm~\ref{alg1} shows our algorithm realized by modifying OCO~\cite{OCO} to bound the worst-class error. 
In each round $t \in [T]$, the weak learning algorithm $\mathcal{A}$ returns $h_t$ by minimizing the error on the distribution with class-weight $\boldsymbol{w}$ as shown in Eq.~\eqref{align:weighted_err}.
The class-weighting algorithm $\mathcal{B}$ gives weights such that the weighted loss of $h_t$ will be large.
The goal of $\mathcal{B}$ is to minimize the regret, 
\begin{align}
&\mathrm{Regret}(\mathcal{B})= \sum_{t=1}^T \boldsymbol{w}_t \cdot \boldsymbol{r}_t - \min_{\boldsymbol{w} \in \Delta_K}\sum_{t=1}^T \boldsymbol{w} \cdot \boldsymbol{r}_t \\ \nonumber  
&= \sum_{t=1}^T\boldsymbol{w}_t \cdot \lparen \boldsymbol{1}-\boldsymbol{\ell}^\theta(h_t) \rparen - \min_{\boldsymbol{w} \in \Delta_K}\sum_{t=1}^T \boldsymbol{w} \cdot \lparen \boldsymbol{1}-\boldsymbol{\ell}^\theta(h_t) \rparen,
\end{align}
when the feedback vector $\boldsymbol{r}_t$ of $\mathcal{B}$ of Eq.~\eqref{align:regret} is defined as the negation of $\boldsymbol{\ell}^\theta (h_t)=({\ell}_1^\theta (h_t),\ldots, {\ell}_K^\theta (h_t))$ as shown in Eq.~\eqref{align:feedback}.
By considering the feedback vector $\boldsymbol{r}_t \in \{0,1\}^K$ as Eq.~\eqref{align:feedback}, 
 $\mathcal{B}$ gives the large weights to the classes with $\ell_k^\theta(h_t)=1$.
As a class-weighting algorithm $\mathcal{B}$, we simply employ the Hedge algorithm~\cite{adaboost}. 
The Hedge algorithm inputs a feedback vector $\boldsymbol{r}_t$  and updates
$\boldsymbol{w}_t=(w_{t,1}, \ldots, w_{t,K}) \in \Delta_K$ according to Eq.~\eqref{align:hedge}. It is known that this algorithm achieves a small regret when one sets $\eta = \sqrt{{8(\ln{n})}/{T}}$~\cite{OCO}.
After the boosting algorithm repeatedly obtains $h_t$ for the corresponding weighted sample with sufficient $T$ rounds, it returns an ensemble hypothesis $\bar{h}$ such that $\ell_k^\theta=0$ for any $k \in [K]$, as proved in the following section.
\par

\noindent\textbf{Difference from standard boosting algorithms}: 
Although standard boosting algorithms such as~\cite{OCO,adaboost} use instance-weighting, our algorithm uses class-weighting $\boldsymbol{w}_t=(w_{t,1},\ldots,w_{t,K})$. In other words, the instances in the same class $k$ share the same weights $w_{t,k}$. This is because the zero-one loss is class-wise but not instance-wise. 
It should be noted that the class-wise weights allow our boosting algorithm to easily accomplish multi-class classification tasks that are difficult for the standard boosting algorithms.

\section{Theoretical analysis}
\subsection{Guarantee of training worst-class error}
\label{subsec:training_theo}
We show that our boosting algorithm has a theoretical guarantee on the training worst-class error. Standard boosting algorithms guarantee that the training average error becomes zero after $T$ rounds.
In contrast, our boosting guarantees that the sum of zero-one losses defined as 
Eq.~\eqref{align:zeroone_train_err} over $K$ classes becomes zero after $T$ rounds.
More formally, we have the following theorem. 
\begin{theo}
\label{theo:trerr_bound}
 Assume that $H$ is $(\gamma, \delta/T, \theta)$-weak learnable
 and the regret of the instance-weighting algorithm $\mathcal{B}$ is bounded by $\frac{\gamma T}{2}$.
    Then, Algorithm~\ref{alg1} returns a hypothesis $\bar{h}$ which verifies the following with probability at least $1-\delta$:
    \begin{align}
    \label{align:theorem_ours}
        {\hat{U}}_{{S}}(\bar{h}) < 1 - \theta.
    \end{align}
\end{theo}\noindent
We can prove it with a similar argument of the proof of Theorem~10.2 in~\cite{OCO}. 
\begin{proof}
For any weak hypothesis $h_t \in \C{H}$ obtained by a weak learning algorithm in each round $t \in [T]$, we have
$\bb{P}\left[\bm{w}_t \cdot \bm{r}_t \le \frac{1}{2}+\gamma \right]
    = \bb{P}\left[\bm{w} \cdot \boldsymbol{\ell}^\theta(h_t) \ge \frac{1}{2}-\gamma \right] 
    \le \frac{\delta}{T}$.
The above can be derived by the assumption of the weak learning algorithm.
From the union bound, we have
    $\bb{P}\left[\frac{1}{T} \sum_{t=1}^T \bm{w}_t \cdot \bm{r}_t \le \frac{1}{2}+\gamma \right] 
    \le \sum_{t=1}^T \frac{\delta}{T} = \delta.$
Combining the these two inequalities, we have
    $\bb{P}\left[\frac{1}{T} \sum_{t=1}^T \bm{w}_t \cdot \bm{r}_t \le \frac{1}{2}+\gamma \right] \le 1-\delta$.
We define $J = \{j \mid \ell_j^\theta(\bar{h}) =1, j \in [K] \}$; 
That is, the ensemble hypothesis $\bar{h}$  has the class-wise error larger than $1-\theta$ for $S_j, j \in [J]$.
Assume that $J \neq \emptyset$ with the class weights $\bm{w}^*$:
    $\sum_{t=1}^{T} \bm{w}_t^* \cdot \bm{r}_t =  \sum_{t=1}^T \frac{1}{|J|} \sum_{j \in J} I\left(\frac{1}{| S_{j} |} \sum_{x \in S_j} I(h_t(x) = j) \ge \theta \right) \nonumber 
    \le \frac{1}{| J |} \sum_{j \in J} \frac{T}{2} = \frac{T}{2}$.
The last inequality assumes that, 
under the assumption that $\ell_k^\theta=1$ for some $k$ and that the majority votes $\bar{h}$, $\ell_k^\theta(h_t)=1$ holds for more than half of $h_1, \ldots, h_T$. 
Then, by using $\bb{P}\left[\frac{1}{T} \sum_{t=1}^T \bm{w}_t \cdot \bm{r}_t \le \frac{1}{2}+\gamma \right] \le 1-\delta$ and 
$\sum_{t=1}^{T} \bm{w}_t^* \cdot \bm{r}_t \leq \frac{T}{2}$, we have
    $\frac{1}{2} + \gamma \le \frac{1}{T} \sum_{t=1}^T \bm{w}_t \cdot \bm{r}_t 
    \le \frac{1}{T} \sum_{t=1}^T \bm{w}_t^* \cdot \bm{r}_t + \frac{1}{T} \mathrm{Regret}_T(\mathcal{B}) 
    \le \frac{1}{2} + \frac{\gamma}{2}.$
This implies that such a weight $\bm{w}^*$ cannot exist, and thus
$\ell_k^\theta(\bar{h}) = 0$ for any $k$.
\end{proof}

\subsection{Generalization bound}
\label{subsec:gen_bound}
Here, we provide the generalization worst-class error bound of our boosting algorithm.
First, we define the Rademacher complexity.
\begin{defi}
[Rademacher complexity~\cite{Bartlett:2003:RGC}]
Let $H$ be a hypothesis set. For a sample $S$,
the Rademacher complexity of 
${H}$ w.r.t. $S$ is defined as
\begin{align}
     \Rdm_S({H})=\frac{1}{n}\mathbb{E}_{\boldsymbol{\sigma}}\left[
\sup_{{h} \in {H}}\sum_{i=1}^n \sigma_i {h}(x_i)
 \right],
\end{align}
where $\boldsymbol{\sigma} \in \{-1,1\}^n$ and each $\sigma_i$ is an independent uniform random variable taking values in $\{-1,+1\}$.
\end{defi}\noindent

It is known that the gap between the generalization and training error can be bounded using $\Rdm_S({H})$~\cite{mohri2018foundations}.

We assume that for a sample $S$ of size $n$ and hypothesis set $H$, there exists $C_{H,S}>0$ such that
$\Rdm_S({H}) \leq \frac{C_{H,S}}{\sqrt{n}}.$
Moreover, we assume that the upper bound of $\Rdm_{S'}({H})$ is less than the upper bound of $\Rdm_S({H})$ for any size $n'$ of $S'$ and any size $n~(<n')$ of sample $S$.
In general, $C_{H, S}$ depends on the richness of the hypothesis set and the given data conditions, e.g., the number of classes and dimensions. 
For example, if $H$ is a set of linear or kernel-based hypotheses, $C_{H,S} = O(\log n^2K)$ and $\Rdm_S({H}) = O\left(\log n^2K/\sqrt{n}\right)$~\cite{suehiro2022simplified}, and thus, it satisfies the assumption of the upper bounds between $\Rdm_{S'}({H})$ and $\Rdm_S({H})$.
Moreover, if we use DNN as a hypothesis set, it is known that $C_{H,S}$ is properly bounded by weight decay~\cite{golowich2018size}.
\par
We derive the following generalization worst-class bound.
\begin{theo}
\label{theo:generr_bound}
Let $H$ be a hypothesis set. Let $\bar{h} \in \bar{H}$ be an ensemble hypothesis over $H$ which is returned by Algorithm~\ref{alg1}, and let $k^* = \arg\max_{k \in [K]}{R}_{{D}_k}(\bar{h})$.
The following holds with probability at least $1-\delta$:
    \begin{align}
    U_D(\bar{h}) \leq 1-\theta + \frac{2C_{H,{S_{k^*}}}}{\sqrt{n_{k^*}}} + 3\sqrt{\frac{\log \frac{2}{\delta}}{2n_{k^*}}}.    
    \end{align}
\end{theo} \noindent
\begin{proof}
We introduce the generalization bound for ensemble hypotheses.
Let $\bar{H}=\{\sum_{t=1}^T a_t I(h_t(x)=y) | h_t \in H\}$ be a hypothesis set\footnote{The majority vote is a special case of a convex combination (see  \cite{zantedeschi2021learning}).}.
Based on Lemma 7.4 and Theorem 3.5 in~\cite{mohri2018foundations}, the following holds with probability at least $1-\delta$ that for all $\bar{h} \in \bar{H}$:
$R_{D}(\bar{h}) \leq \hat{R}_{S}(\bar{h}) + 2\Rdm_S({H}) + 3\sqrt{{\log ({2}/{\delta})}/{2n}}$.
Let us focus on the class-wise error. If we divide $D$ into $D_1, \ldots, D_K$
and $S$ into $S_1, \ldots, S_K$, for any $k \in [K]$, the following holds with the probability at least $1-\delta$:
$    R_{D_k}(\bar{h}) \leq \hat{R}_{S_k}(h) + 2\Rdm_{S_k}({H}) + 3\sqrt{{\log ({2}/{\delta})}/{2n_k}},$
Using the assumption that the upper bound of $\Rdm_{S_k}(h)$ is less than the upper bound of $\Rdm_{S_{k*}}$ for any $k \in [K]$ and Theorem~1, we have
    $U_D(\bar{h})=R_{D_{k^*}}(\bar{h}) \leq 1-\theta + 2\Rdm_{S_{k^*}}({H}) + 3\sqrt{{\log ({2}/{\delta})}/{2n_{k^*}}}.$
Using the assumption that $\Rdm_S({H}) \leq {C_{H,S}}/{\sqrt{n}}$,
we obtain the theorem. 
 \end{proof}
Additionally, we derive the following generalization bound without using a (possibly) unknown parameter ${k^*}$.
\begin{coro}
Let $\bar{h} \in \bar{H}$ be a hypothesis that is returned by Algorithm~\ref{alg1}. Let $\tilde{k} =\arg\min_{k \in [K]}n_k$.
The following holds with a probability of at least $1-\delta$:
\begin{align}
        U_D(\bar{h}) \leq 1-\theta +         
        \frac{2C_{H,S_{\tilde{k}}}}{\sqrt{n_{\tilde{k}}}}           
        + 3\sqrt{\frac{\log \frac{2}{\delta}}{2n_{\tilde{k}}}}.
\end{align}
\end{coro}\noindent
The above can be easily derived by recognizing the fact that $n_{\tilde{k}}=\min_{k\in [K]}n_{k} \leq n_{k^*}$ and using the assumption between the upper bounds of $\Rdm_{S_k}({H}) $ and $\Rdm_{S_{\tilde{k}}}({H})$.
\par
An important notion of the above theorem and corollary is that the worst-class generalization error
mainly depends on $n_{k^*}$ (i.e., the sample size of the hardest class in hindsight) or $n_{\tilde{k}}$ (i.e., the sample size of the smallest class).
Past literature often said that the performance of classification methods 
depends on the imbalance ratio, which is defined as $\rho=\max_k{n_k}/\min_k{n_k}$ (e.g.,~\cite{buda2018systematic,IB}). 
However, our generalization bound only depends on $n_{\tilde{k}}$, not the imbalance ratio.
This means that, even for a highly imbalanced training sample, our boosting algorithm achieves a good generalization performance if there are a sufficiently large number of training instances over any class.

\section{Experiments} \label{sec:exp}
First, we validate that our method behaves as expected using artificial toy datasets. Next, we compare our method with existing approaches on real datasets to demonstrate its effectiveness for both balanced and imbalanced datasets\footnote{The code is available at \url{https://github.com/saito-yuya/Bounding-the-Worst-class-Error-A-Boosting-Approach.git.}}.
\subsection{Experimental settings}

\begin{figure}[t]
\centering
    \centering
    \includegraphics[keepaspectratio, width=1.0\linewidth]{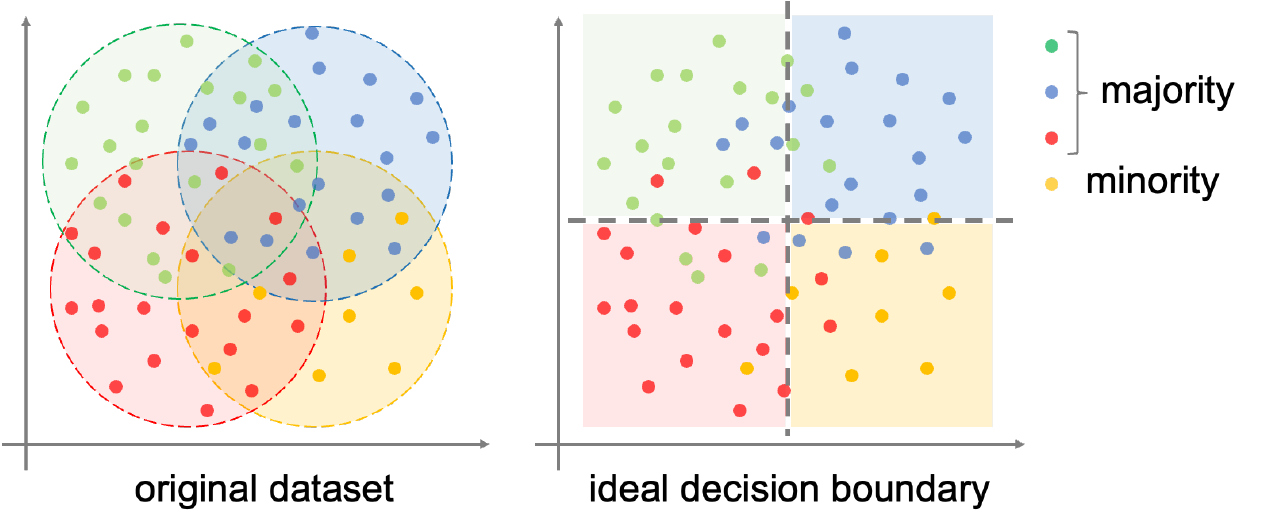}
    
    \caption{Illustration of the artificial imbalanced dataset (left) and its ideal decision boundary without bias in the class-wise error (right). }
    \label{fig:imbalanced_data}
\end{figure}

\textbf{Datasets}:
We use two artificial and two real datasets. For the artificial datasets, we created training and test samples five times and computed the average scores.
\begin{itemize}
    \item A balanced artificial dataset: It consists of 5 (classes) $\times$ 100 (instances)  two-dimensional instances for training and 5 $\times$ 100,000 for tests. Fig.~\ref{fig:balance_toydata} shows a part of the training sample; the instances within each (square) class region are generated randomly and uniformly.
    \item Imbalanced artificial datasets: Each of them consists of 4 (classes) two-dimensional instances. Fig.~\ref{fig:imbalanced_data} (left) shows a part of the training sample; the instances within each (circular) class region are generated randomly and uniformly. The yellow class is the minority and has $n_{\tilde{k}}$ instances; the other three classes have $10n_{\tilde{k}}$ instances. (This means we fixed the imbalance ratio $\rho=10$.) In the experiment, we examined $n_{\tilde{k}}\in \{10, 50, 100\}.$  The test sample sizes are 10,000 for the minority class and 100,000 for the other classes.
    \item Balanced real datasets: The CIFAR-10, CIFAR-100, and Tiny ImageNet datasets. The original training sample was divided into training and validation samples with a ratio of $7:3$.
    \item Imbalanced real datasets: Imbalanced CIFAR-10~\cite{IB,CB,LDAM}, EMNIST (ByClass)~\cite{cohen2017emnist}, and TissueMNIST~\cite{yang2023medmnist}. The imbalance ratios $\rho$ of these datasets are 10, 20.24, and 9.05, respectively.   
    For Imbalanced CIFAR-10, we create the test sample with the same imbalance ratio as the training sample. For EMNIST, we divided the original training sample into training and validation samples with a ratio of $7:3$ For TissueMNIST, we used the official training, validation, and test splits.
\end{itemize} 

\noindent\textbf{DNNs as weak learners}: 
We use DNNs as weak learners in our boosting algorithm. At each round $t$, we internally train a DNN by multiple epochs to find $h_t$ with the class-weight $\boldsymbol{w}_t$. A practical issue is that DNN is a rich hypothesis and easily overfits to training sample with sufficient epochs. In fact, our boosting algorithm often stops at the first round ($T=1$) by achieving $r_{k,1}(h_1)=1$ for all $k \in [K]$. Therefore, we 
terminate the training process of individual DNNs before overfitting. More specifically, 
we stop the training when its performance on the weighted sample (i.e., $\sum_{k=1}^K w_{t,k}\hat{R}_{S_k}(h)$) becomes larger than $1/2 + \gamma$.
\par  

Other settings for training the DNN for the $t$-th round are as follows. The loss function is the cross-entropy with class-weight $\boldsymbol{w}_t$. The maximum number of rounds $T$ was determined in accordance with the definition given for the weak-learnability. (See Definition~\ref{def:weak-learnability} and Theorem~\ref{theo:trerr_bound}.)
If the weak learning algorithm could not find a hypothesis $h_t$
satisfying the weak learnability definition (that is, we cannot obtain a DNN with sufficient performance on the weighted sample over large epochs), we stopped the round and obtained the ensemble of $h_1, \ldots, h_{t-1}$.

\smallskip\noindent\textbf{Hyperparameters}:
The hyperparameter $\gamma$ needs to be determined according to the number of classes $K$. 
We, therefore, prepared several candidate values for $\gamma$ according to $K$ and have conducted several preliminary experiments; however, their results showed that the value of $\gamma$ is not very sensitive to the results. Consequently, $\gamma$ was set at about $0.1$ for the artificial datasets and $0.3$ for the real datasets.
The hyperparameter $\theta$ is determined for each dataset. For the artificial datasets, we know their sample distribution and thus can predetermine the appropriate $\theta$. For the real datasets, we use the validation set to fix $\theta$. 
 

\smallskip\noindent\textbf{Comparative methods}:
\begin{itemize}
    \item CE: DNN trained with the cross-entropy loss.
    \item wCE: CE with fixed class weights $1/n_k$.
    \item Focal: DNN trained with the focal loss~\cite{focal_loss}.
    \item CB: DNN trained with the class balanced loss~\cite{CB}.
    \item IB: DNN trained with the influence-balanced loss~\cite{IB}.
    \item IB+CB: DNN trained with IB and CB.
    \item IB+Focal: DNN trained with IB and focal loss.
    \item LA: DNN with logit adjustment(LA)~\cite{LA} based on the label frequencies.
    \item CDT: DNN using class-dependent temperatures (CDT)~\cite{CDT} for increasing the performance on minor class considering feature deviation.
    \item VS: DNN trained with Vector-Scaling (VS) loss~\cite{VSloss}.
    \item Boosting: OCO~\cite{OCO} for minimizing the average error.
    \item Naive: DNN trained to minimize the training worst-class cross-entropy loss.
    More precisely, the DNN minimized $\max_{k \in [K]}\frac{1}{n_k}\sum_{x \in S_k} L_k(h,x)$, where $L_k$ is the cross-entropy loss corresponding to class $k$. 
    \item 0.5-DNN: By setting $\gamma=0.5$ at Eq.~\eqref{def:weak-learnability}, our algorithm finds a single DNN $h_1$ that satisfies Eq.~\eqref{align:objective}. 
\end{itemize}
Among them, Focal~\cite{focal_loss}, CB~\cite{CB},IB~\cite{IB}, and VS~\cite{VSloss} are the state-of-the-art approaches to deal with class imbalance.

Due to the high computational cost of the weak learning of ``Boosting'', we stopped training after one week in our environment (PyTorch, Xeon Gold 6338, A100 80GB).

\begin{figure}[t]
 \centering
 \begin{minipage}{0.48\linewidth}
    \includegraphics[keepaspectratio, width=\linewidth]{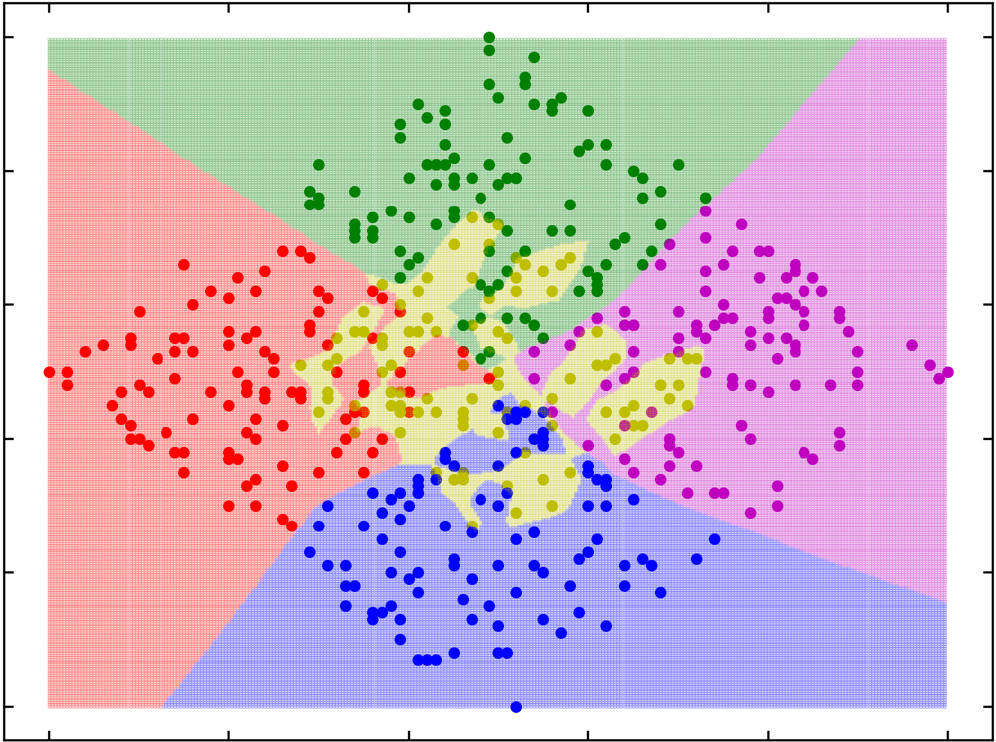}
    \subcaption{CE}
 \end{minipage}
  \begin{minipage}{0.48\linewidth}
    \includegraphics[keepaspectratio, width=\linewidth]{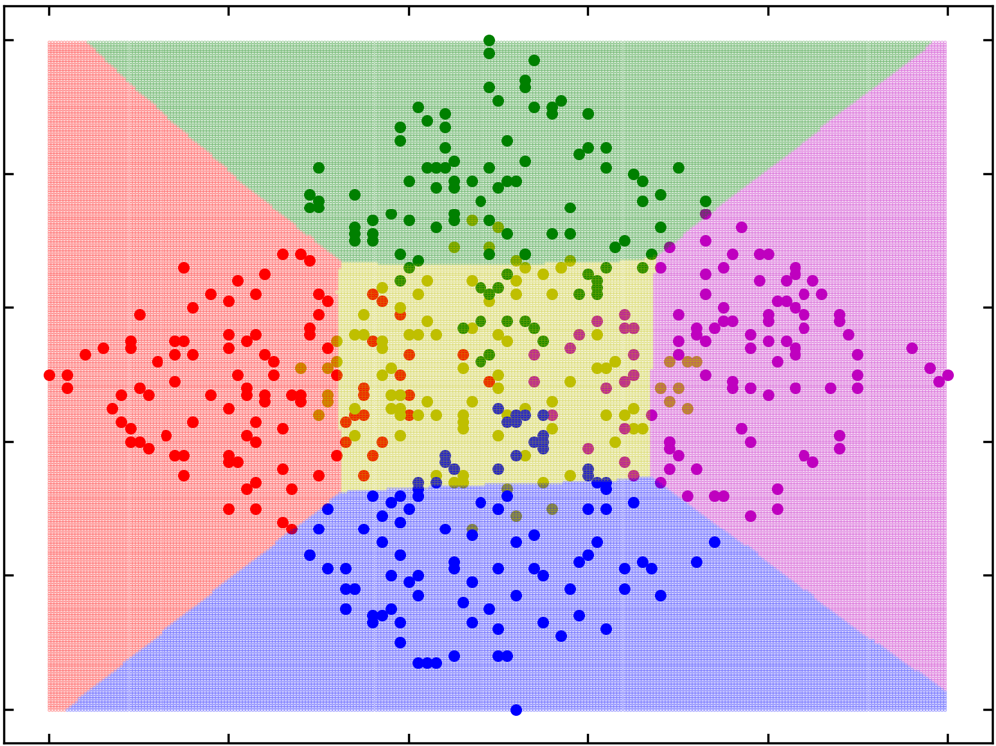}
    \subcaption{Ours}
 \end{minipage}
 
 \caption{Visualization of the decision boundaries on the balanced artificial dataset.}
 \label{fig:toy_balance_area}
\end{figure}

\begin{table}[t]
\caption{Classification errors [\%] on artificial balanced dataset.}
\label{tb:toytable}
\centering
\begin{tabular}{lccccccc}
\hline 
\multirow{2}{*}{metrics} & \multicolumn{5}{c}{{class-wise training error}} &\multicolumn{2}{c}{{test error}} \\
\cline{2-8} 
          &   1     & 2     & 3    & 4    & 5    &  Worst   & Avg. \\ \hline \hline
CE             &3.2  &13.2 &4.0  &2.6  &2.0 &  52.2          & 23.4    \\
Boosting~\cite{OCO}      &1.4  &5.4  &1.4  &1.4  &0.6 &  49.6          & 23.3    \\
0.5-DNN        &13.2  &20.0  &15.6  &18.2  &15.4 &  32.0 & \textbf{22.7}    \\
Naive          &20.8 &17.2 &26.4 &20.4 &18.2&  28.5          & 23.8    \\ 
\rowcolor{gray!15}
Ours           &20.6 &24.0 &19.8 &20.0  &20.2 &  \textbf{28.1} &  23.9 \\ \hline
\end{tabular}
\end{table}

\begin{figure}[t]
\centering
    \centering
    \includegraphics[keepaspectratio, width=1.0\linewidth]{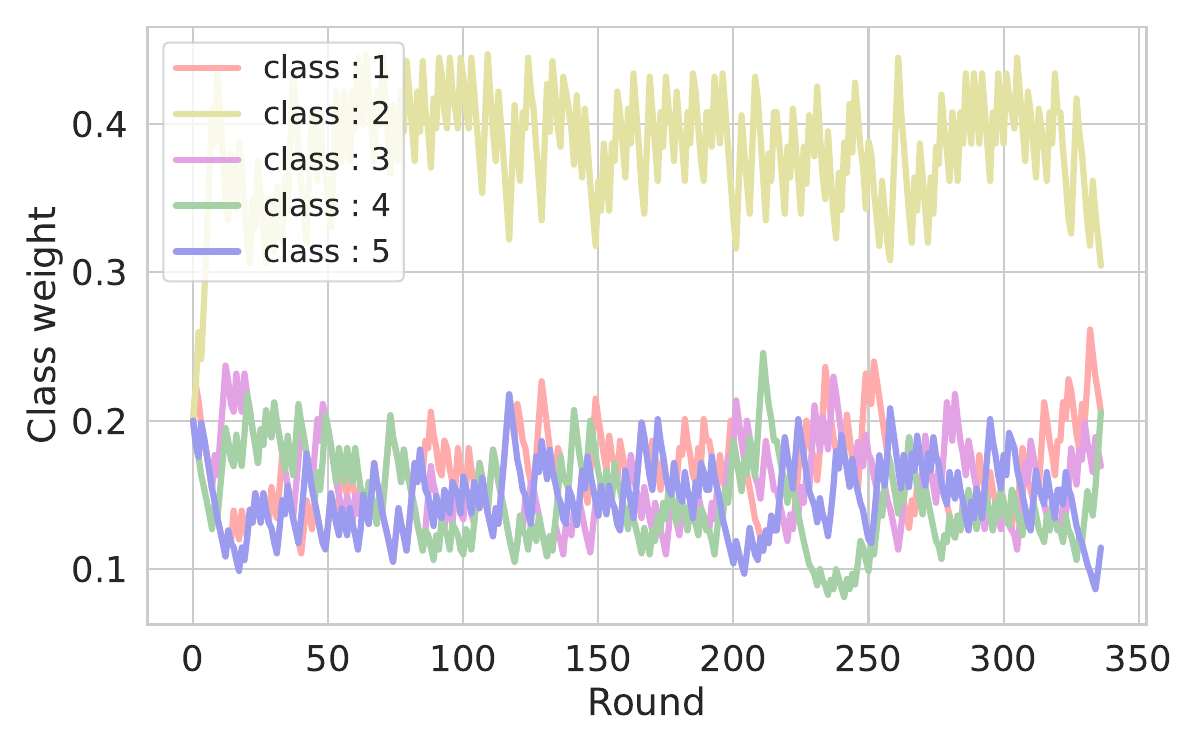}
    \caption{Change in class weights of our boosting on a balanced artificial dataset.}
    \label{fig:weight_curve}
\end{figure}

\subsection{Results on artificial datasets}
\textbf{Balanced artificial dataset}:
The training class-wise errors, worst-class, and average test errors of the balanced artificial datasets are shown in Table~\ref{tb:toytable}. Here, the hyperparameter $\theta$ was set at $0.75$. For this balanced problem, we did not show the results of the methods for imbalanced cases.
This table shows the following facts. First, among all methods, ours successfully achieved the smallest worst-class test error. Second, our boosting was able to bound the training worst-class error by $1-\theta=25\%$ for all five classes. Third, the other methods achieved larger test worst-class errors, whereas they have slightly better average errors than ours. This means that the other methods sacrifice some classes.
\par 


The decision boundaries of CE and ours are shown in Fig.~\ref{fig:toy_balance_area}.
We can see that the decision boundaries of CE seem to be overfitted.
By contrast, the decision boundary of our boosting indicates that it avoided the bias of the class-wise error and overfitting.
\par  
Fig.~\ref{fig:weight_curve} shows the changes in the class weights $\boldsymbol{w}_t=(w_{t,1},\ldots, w_{t,K})$ over the rounds. We can see that our algorithm gave the larger weight to the yellow class (i.e., the minority class). This result shows that our boosting effectively bound the worst-class error and verifies the theoretical results.
\par  

\begin{table}[t]
    \caption{Worst-class and average test errors [\%] on artificial imbalanced dataset.}
    \label{tb:toytable2}
    \begin{tabular}{lcccccc}
    \hline
    {$n_{\tilde{k}}=\min n_k$} & \multicolumn{2}{c}{10}                                  &  \multicolumn{2}{c}{50}                                  &  \multicolumn{2}{c}{100}                                 \\ \hline 
    \multicolumn{1}{l}{metrics}   & {Worst} & {Avg.} &   {Worst} & {Avg.} &  {Worst} & {Avg.} \\ \hline \hline
    CE                  & {71.0}  & {30.8} &  {68.0}  & \textbf{29.2}        &  {65.1}  & {29.0}    \\
    wCE        & {64.8}  & {32.1}        &  {53.8}  & {33.2}        &  {47.5}  & {33.8}    \\
    Boosting~\cite{OCO}         & {67.6}  & {31.1}        &  {67.8}  & {29.1}        &  {66.4}  & \textbf{28.8}    \\ 
    0.5-DNN  & {67.3}  & {\textbf{30.4}}        &  {63.9}  & {29.9}        &  {57.9}  & 32.1    \\
    Naive               & \textbf{55.5}  & {34.5} &  {52.2}  & {33.4}        &  {51.1}  & {33.8}    \\ 
    \rowcolor{gray!15}
    Ours                & {59.8}  & {31.3}        &  \textbf{49.5}  & {33.2}        &  \textbf{45.5}  & {34.1}    \\ \hline
    \end{tabular}
    
\end{table}

\smallskip\noindent\textbf{Imbalanced artificial dataset}:
Table~\ref{tb:toytable2} shows the worst-class test errors of our boosting and the comparative methods for imbalanced cases. Here, the hyperparameter $\theta$ was set at $0.5$ since this dataset is more difficult than the balanced dataset (especially for the minority class). 
Our boosting achieved the smallest worst-class error except when $n_{\tilde{k}}=10$. 
We can observe that the worst-class error decreased with increasing training sample size of the smallest minority class (i.e., $\arg\min_{k \in [K]}n_k$). 
This result verifies Theorem~\ref{theo:generr_bound} and shows that our $\theta$-bounded approach is effective for imbalanced data with sufficiently large $\min_{k \in [K]}n_k$.
\par
\begin{figure*}[t]
 \centering
 \begin{minipage}[t]{0.24\linewidth}
    \includegraphics[keepaspectratio, width=\linewidth]{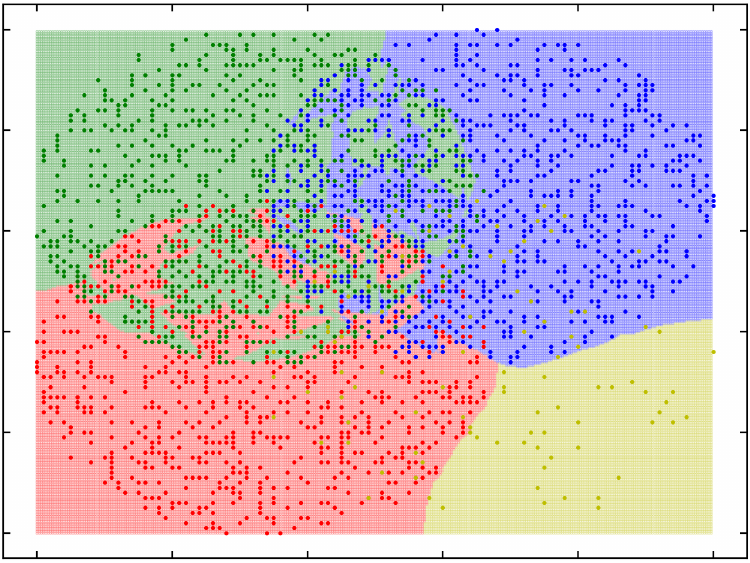}
    \subcaption{Boosting with \\ $\min_{k \in [K]}{n_k}=100$} 
 \end{minipage}
 \hspace{0.04\columnwidth}
  \begin{minipage}[t]{0.72\linewidth}
    \includegraphics[keepaspectratio, width=\linewidth]{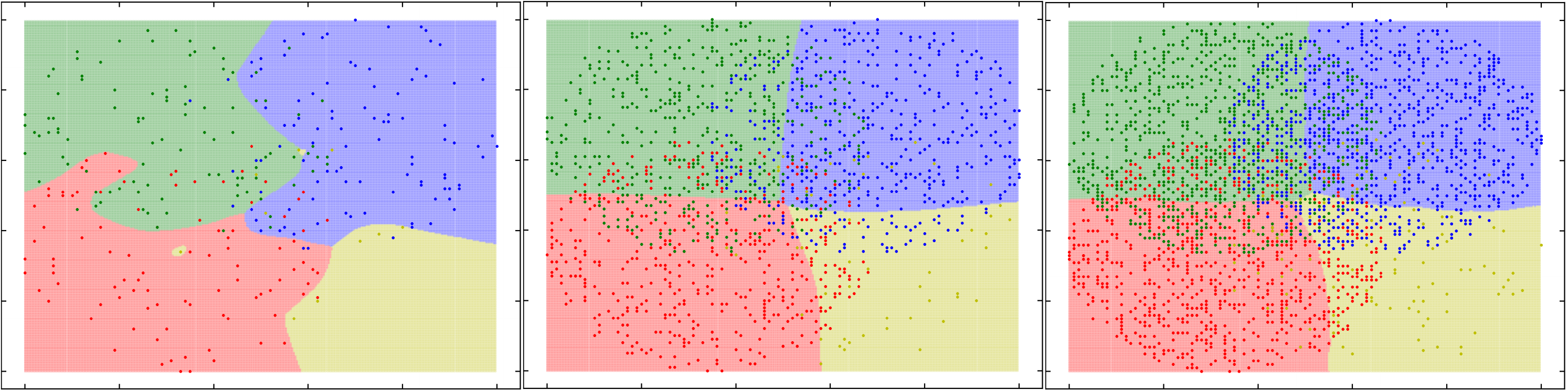}
    \subcaption{Ours with $\min_{k \in [K]}{n_k}=10, 50, 100$}
 \end{minipage}
 \caption{Visualization of the decision boundaries on the imbalanced artificial dataset.}
\label{fig:all_class_area_imbalance}
\end{figure*}

The decision boundaries of our boosting and  Boosting are shown in Figure~\ref{fig:all_class_area_imbalance}. 
We can see that
our boosting improved the decision boundary with increasing 
$\min_{k \in [K]}{n_k}$ even though the imbalance ratio $\rho$ was fixed.
Boosting overfitted to the training sample even when $\min_{k \in [K]}{n_k}=100$.
This results support for Theorem~\ref{theo:generr_bound}.

\begin{table}[t]
\caption{Worst-class and average test errors [\%] for balanced real datasets.}
\label{tab:balanced_real_datasets}
\centering
\begin{tabular}{lcccccc}
\hline
                 {datasets}&  \multicolumn{2}{c}{{CIFAR-10}} &  \multicolumn{2}{c}{{CIFAR-100}} &  \multicolumn{2}{c}{{Tiny ImageNet}} \\ \hline 
\multicolumn{1}{l}{{metrics}} &   {Worst}        & {Avg.}       &  {Worst}        & {Avg.}        &               {Worst}       & {Avg.}          \\ \hline \hline
CE                 & 30.9                  & 14.9                                         & 78.0                  & 46.7                                          & 84.0                    & 42.8                      \\
Boosting~\cite{OCO}         & 26.2                  & 13.1                                         & 66.0                  & \textbf{35.4}                               & 72.0                    & 37.2                      \\ 
Naive              & 23.0                  & 17.2                                         & 87.0                  & 54.9                                          & 100.0                   & 81.7                      \\
Focal~\cite{focal_loss}                & 25.6                  & 14.2                                         & 80.0                  & 46.5                                          & 82.0                    & 43.0                      \\
IB~\cite{IB}                & 28.1                  & 13.7                                         & 92.0                  & 47.8                                          & 90.0                    & 41.7                      \\
IB+Focal~\cite{IB}             & 25.6                  & 13.8                                         & 86.0                  & 46.7                                          & 90.0                    & 41.7                      \\
LA~\cite{LA}                  &  31.4                 & 14.2                                        & 80.0                  & 46.5                                           & 86.0                    & 42.4                     \\
CDT~\cite{CDT}                 & 35.0                    & 13.8                                       & 82.0                  & 46.2                                          & 86.0                   & 42.6                      \\
VS~\cite{VSloss}                  & 27.1                   & 13.8                                       & 84.0                  & 46.7                                        & 90.0                    & 42.4                    \\
\rowcolor{gray!15}
Ours          & \textbf{18.9}         & \textbf{9.6}                                & \textbf{62.0}          & 44.5                                          & \textbf{68.0}           & \textbf{36.1}             \\ \hline
\end{tabular}
\end{table}


\subsection{Results on real datasets}
\noindent\textbf{Balanced real datasets}:
We show the worst-class test errors and average test errors in Table~\ref{tab:balanced_real_datasets}. Here, $\theta$ was determined among nine candidates $\{0.1, 0.2, \ldots, 0.9\}$ by using the smallest validation worst-class error.
We can see that our boosting algorithm successfully outperformed the comparative methods for all three datasets in their worst-class errors; the comparative methods sacrifice some classes in the datasets. 
\par 
Our boosting algorithm also achieved the smallest average error for CIFAR-10 and Tiny ImageNet. Since we simply used the worst-class error for the validation, we cannot conclude that bounding the worst-class error is always useful for achieving smaller average errors; however, our result indicates that bounding the worst-class error does not always have a negative impact on average errors. 
\par 

\smallskip\noindent\textbf{Imbalanced real datasets}:
Table~\ref{tab:imbalanced_real_datasets} shows the worst-class test errors and average test errors
on the three datasets. Here, the hyperparameter $\theta$ was determined for each dataset by the same procedure as the balanced real datasets.
Like the balanced case, our boosting algorithm performed the best on all three datasets in its worst-class error. Recall that in practical classification tasks, especially medical image classification tasks, sacrificing certain classes should be avoided. Among the three datasets, 
TissueMNIST is a medical image dataset. Even using special loss functions that make DNNs robust to 
class imbalance, the worst-class error on TissueMNIST becomes really high (such as 76.5\% error).
In contrast, ours can reduce the increase in the worst-case class error. This result supports the practical usefulness of the proposed algorithm, although the CE and state-of-the-art methods for imbalanced cases achieved smaller average errors than ours on EMNIST and TissueMNIST.
\par 

\begin{figure}[t]
 \centering
 \begin{minipage}{0.49\linewidth}
    \includegraphics[keepaspectratio, width=\linewidth]{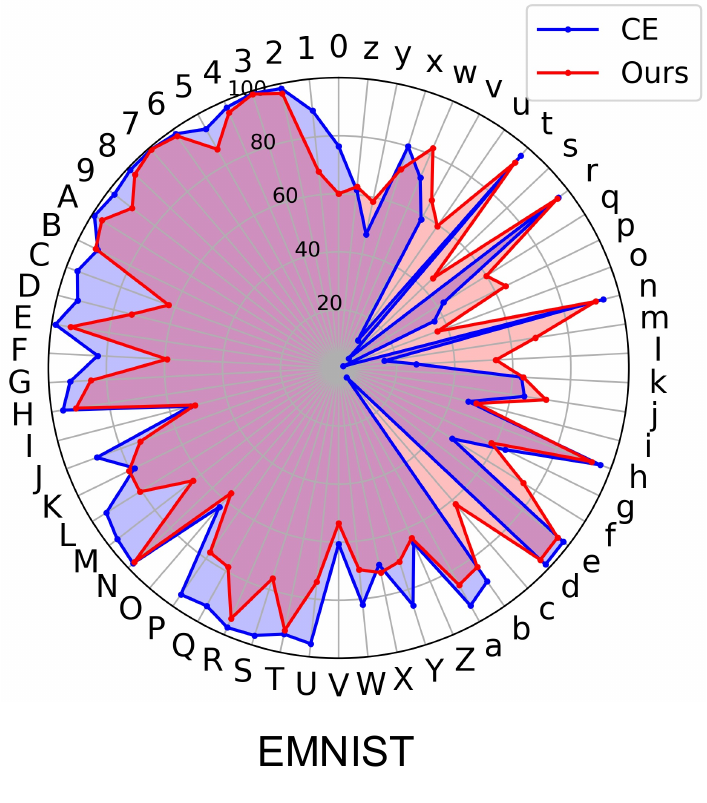}
 \end{minipage}
 \begin{minipage}{0.49\linewidth}
    \includegraphics[keepaspectratio, width=\linewidth]{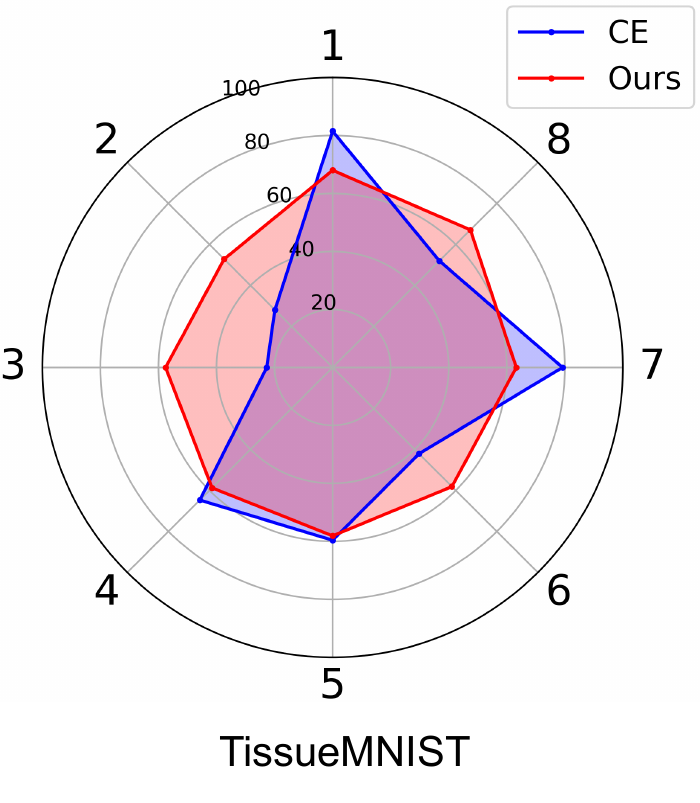}
 \end{minipage}
 \caption{Radar charts: Class-wise test {\em accuracy} [\%] on EMNIST and TissueMNIST.}
 \label{fig:radar_chart}
\end{figure}

Fig.~\ref{fig:radar_chart} shows the class-wise test accuracies of our boosting algorithm and CE on EMNIST and TisuueMNIST. The results with CE clearly indicate multiple classes are sacrificed to achieve smaller average errors. In contrast, ours shows less biased class-wise errors than CE.
\par  

\begin{table}[t]
\caption{Worst-class and average test errors [\%] for imbalanced real datasets.}
\label{tab:imbalanced_real_datasets}
\centering
\begin{tabular}{lcccccc}
\hline
\multicolumn{1}{l}{datasets}               & \multicolumn{2}{c}{Imb. CIFAR-10} & \multicolumn{2}{c}{EMNIST} & \multicolumn{2}{c}{TissueMNIST} \\ \hline
\multicolumn{1}{l}{$\rho$}                & \multicolumn{2}{c}{10}                  & \multicolumn{2}{c}{20.24}       & \multicolumn{2}{c}{9.05}   \\ \hline
\multicolumn{1}{l}{$\min_{k \in [K]}n_k$} & \multicolumn{2}{c}{350}                 & \multicolumn{2}{c}{1327}       & \multicolumn{2}{c}{5866}   \\ \hline
\multicolumn{1}{l}{metrics}         & Worst               & Avg.              & Worst        & Avg.        & Worst        & Avg.        \\ \hline
CE                                        & 41.0                & 21.0              & 98.3         &\textbf{12.4}& 73.0         & 33.9        \\
wCE                              & 47.0                & 21.2              & 89.3         & 16.2        & 69.6         & 34.4        \\
Boosting~\cite{OCO}                                & 47.0                & 19.4              & 97.9         & 12.9        & 80.8         & 34.3          \\
Naive                                     & 46.2                & 35.8              & 73.9         & 38.1        & 50.5         & 41.3        \\
Focal~\cite{focal_loss}                                       & 36.2                & 21.0              & 96.8         & 12.7        & 71.0         &\textbf{33.2}\\
CB~\cite{CB}                                          & 39.9                & 20.7              & 90.6         & 13.7        & 66.9         & 34.4        \\
IB~\cite{IB}                                          & 46.4                & 19.6              & 98.1         & 16.4        & 86.2         & 34.9        \\
IB+CB~\cite{IB}                                       & 47.8                & 20.2              & 97.4         & 12.5        & 86.2         & 34.9        \\
IB+Focal~\cite{IB}                                    & 46.8                & 20.0              & 97.4         & 12.5        & 86.2         & 34.9        \\ 
LA~\cite{LA}                                          & 43.0                & 19.9              & 75.1         & 21.2        & 63.7         & 42.2        \\
CDT~\cite{CDT}                                        & 36.9                 & 20.4              & 74.6         & 22.7        & 61.5         &  39.2       \\
VS~\cite{VSloss}                                         & 31.7                 & 16.5              & 71.9         & 21.2        & 76.5         &   40.3      \\
\rowcolor{gray!15}
Ours                                      & \textbf{29.0}       & \textbf{14.9}     &\textbf{63.7} & 17.5        &\textbf{47.2} & 36.4        \\ \hline
\end{tabular}
\end{table}

\subsection{Discussion on hyperparameters}
\label{subsec:sensitivity}
Since the performance of our algorithm highly depends on $\theta$,
we introduce the following three approaches for obtaining a good $\theta$.
The first approach is, as in the experiments, preparing the candidates of $\theta$ and choosing the best one using the validation set. 
The second is to prepare the candidates of $\theta$ according to some baseline methods. This is a reasonable scenario in practice if we have already observed the worst-class error of some learning method and want to improve from it. For example, if we consider the improvement from CE in Table~\ref{tab:balanced_real_datasets} on CIFAR-10, we need to set $\theta\geq 1-0.309$.
The third is to prepare the candidates according to the goal of the task.
For example, in cancer-subtype recognition~\cite{gao2019deepcc}, if any cancer-subtypes should be classified with more than $80$\% accuracy, we should set $\theta \geq 0.8$.
\par
The hyperparameter $\gamma$ was not quite sensitive to the worst-class error performance in our preliminary experiment,
and thus we fixed it for the experiments on the real datasets. 

\section{Conclusion}
We focused on the worst-class error and provided the problem formulation.
To avoid overfitting when using a rich hypothesis class, such as DNN, we considered bounding the worst-class error. We theoretically designed a boosting algorithm specialized for bounding the worst-class error and derived the training and generalization performance.
The experimental results complemented our theoretical results and showed that the boosting algorithm successfully avoided the bias of class-wise errors on artificial/real and balanced/imbalanced datasets.






\clearpage
\bibliographystyle{splncs04}
\bibliography{main}   
\clearpage

\begin{appendices}
\textbf{Note}: This appendix is included only in the arXiv version of the paper and is not part of the published version.

\section{Details of experimental settings}
\subsection{Implementation}
We implemented our boosting and other baselines with Pytorch, and the code of our implementation can be found in \url{https://github.com/saito-yuya/Bounding-the-Worst-class-error-A-Boosting-Approach}. 

\subsection{Datasets}
The detail of the datasets used in our experiments is shown in Table ~\ref{tab:details_dataset}.

\begin{table*}[t]
\begin{center}
\caption{The details of datasets. }
\begin{tabular}{|c|c|c|c|c|}
\hline
Datasets              & \#Training        & \#Validation & \#Test & \#Class \\ \hline
Artificial balanced   & 500               & -            & 500000 & 5       \\ \hline
Artificial imbalanced & \{310,1550,3100\} & -            & 310000 & 4       \\ \hline
CIFAR-10              & 35000             & 15000        & 10000  & 10      \\ \hline
CIFAR-100             & 35000             & 15000        & 10000  & 100     \\ \hline
Tiny ImageNet         & 70000             & 30000        & 10000  & 200     \\ \hline
Imbalanced CIFAR-10($\rho=10$)   & 14301             & 6130         & 4084   & 10      \\ \hline
EMNIST                & 488552            & 209380       & 116323 & 62      \\ \hline
TissueMNIST           & 165466            & 23640        & 47280  & 8       \\ \hline
\end{tabular}
\label{tab:details_dataset}
\end{center}
\end{table*}

\subsection{DNN structure and optimizer}
We used a multilayer perceptron (MLP) on the artificial datasets and ResNet32 on the CIFAR datasets and ResNet18 on Tiny ImageNet, EMNIST, and TissueMNIST. 
The MLP architecture is shown in Table~\ref{tab:MLP_arch}.
The optimizer was Adam. We set the learning rate for the artificial datasets to 0.01 and used the default setting for the real datasets.
The batch size was 512 for CIFAR-10, CIFAR-100, EMNIST, and TissueMNIST, and 128 for Tiny ImageNet.

\begin{table*}[t]
\begin{center}
\caption{The detail of MLP architecture}
\begin{tabular}{|c|c|}

\hline
Dataset                                             & Artificial \{Balanced, Imbalanced\} datasets                                           \\ \hline
\multicolumn{1}{|c|}{Input}                         & 2                                                             \\ \hline
\multicolumn{1}{|r|}{\multirow{2}{*}{Hidden layer}} & fc1, 1024                                                     \\
\multicolumn{1}{|r|}{}                              & fc2,  512                                                                      \\ \hline
\multicolumn{1}{|c|}{Output}                        & \begin{tabular}[c]{@{}c@{}}fc, \#class\\ softmax\end{tabular} \\ \hline
\end{tabular}
\label{tab:MLP_arch}
\par - ReLU was used as an activation function
\end{center}
\end{table*}

\subsection{Stopping rules of baselines and our boosting}
The maximum epoch was set as $10000$.
For the artificial datasets, we stopped the epochs of CE and Naive if the training loss was not updated in 100 epochs.
For both artificial and real datasets, we stopped the rounds (and epochs) of ``Boosting'' and our boosting if $\bm{w}_t \cdot \bm{r}_t$ is not updated in 1000 epochs of weak learning at $t$.
Since the computation cost of the weak learning of ``Boosting'' is too high, we stopped the round if the total training time reached one week in our computational environment (Pytorch, Intel Xeon Gold 6338 (2.00GHz, 32Core) A100, 80GB GPU memory).

\section{About hyper-parameters of our boosting}
We summarize the values of $\gamma$ and $\theta$ of our boosting in Table~\ref{tab:sets_of_theta}.
For our boosting algorithm, we consider the candidates of $\gamma$ for all $K$-class balanced/imbalanced real datasets as below:
\begin{align}
    \label{align:gamma}
    \gamma =  \frac {\lfloor p \times K \rfloor}{K} - \frac{1}{2} - \epsilon,
\end{align}
where $p \in (0.5, 1.0]$, $\epsilon$ is a small positive value to avoid numerical computation error, in the experiments, we set the $\epsilon = 0.0005$ for all datasets.
At 1st round of our boosting, $1/2-\gamma$ means the rate of the class such that $\ell_k^\theta$ can be 1. Then, we gave reasonable candidates of $\gamma$ depending on the number of classes.
In this work, we consider $p \in \{0.6, 0.7, 0.8, 0.9\}$ (i.e., \{60, 70, 80, 90\}\% of $\ell_k^\theta$ should be 0 at 1st round).
Based on the preliminary experiments, we fixed $p=0.3$ and $p=0.8$ for the artificial and the real datasets, respectively (see also Table~\ref{tab:gamma_sensitivity}).
\par
\paragraph{Sensitivity of $\gamma$}
Table~\ref{tab:gamma_sensitivity} shows the result for CIFAR-10 with $\theta=0.9$. We can see that $\gamma$ was not sensitive to both average and worst class accuracy. 

\begin{table*}[t]
    \caption{Candidates of $\gamma$ and $\theta$ for our boosting. }
    \begin{center}
      \begin{tabular}{l|ccc}
        \hline
        datasets              & $\gamma$ &  & $\theta$            \\ \hline \hline
        artificial balanced   & 0.0995   &  & 0.75                \\
        artificial imbalanced & 0.0995   &  & 0.5                 \\
        balanced CIFAR-10     & 0.2995   &  & \{0.1,0.2,...,0.9\} \\
        balanced CIFAR-100    & 0.2995   &  & \{0.1,0.2,...,0.9\} \\
        balanced TinyImage    & 0.2995   &  & \{0.1,0.2,...,0.9\} \\
        imbalanced CIFAR-10   & 0.2995   &  & \{0.1,0.2,...,0.9\} \\
        EMNIST                & 0.2898   &  & \{0.1,0.2,...,0.9\} \\
        TissueMNIST           & 0.2495   &  & \{0.1,0.2,...,0.9\} \\ \hline
      \end{tabular}
    \end{center}
    \label{tab:sets_of_theta}
\end{table*}

\begin{table*}[h]
    \caption{Effects of $\gamma$ on worst-class and average test errors [\%] on the balanced CIFAR-10 dataset.} 
    \centering
    \begin{tabular}{lcccc}
    \hline
    $\gamma$ & 0.0995 $(p=0.6)$~~  & 0.1995 $(p=0.7)$~~  &  0.2995 $(p=0.8)$~~  & 0.3995 $(p=0.9)$~~ \\ \hline \hline
    Avg.     & 10.4 & 10.1 &  9.6 & 9.4   \\ 
    Worst    & 19.9 & 20.2 &  18.9 & 19.7   \\\hline
    \end{tabular}
    \label{tab:gamma_sensitivity}
\end{table*}

\par
\paragraph{Effect of $\theta$}
We investigated the effect of $\theta$ on several datasets.
As shown in Table~\ref{tab:theta_balance_sensitivity} and \ref{tab:theta_imbalance_sensitivity}, $\theta$ largely influences the worst-class test error for balanced and imbalanced real datasets.
However, as written in the paper, we successfully tuned $\theta$ by the validation set.


\begin{figure}[t]
 \centering
    \includegraphics[keepaspectratio, width=\linewidth]{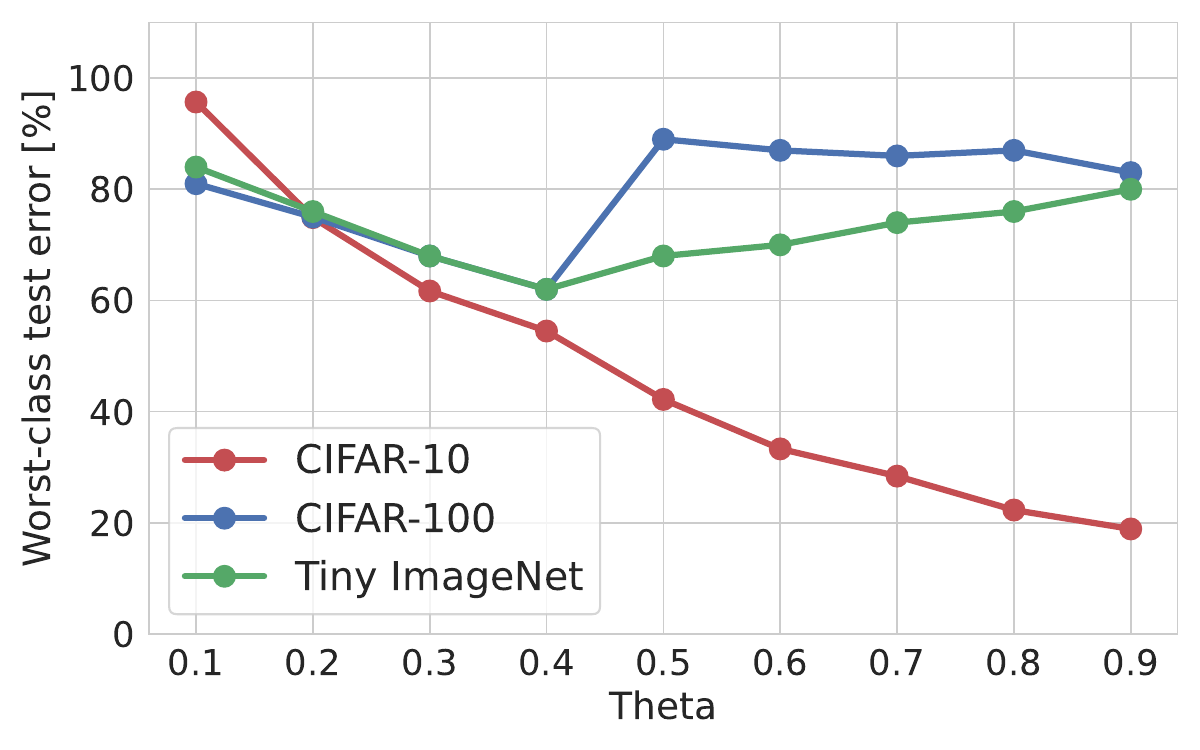}
    \caption{Balanced:CIFAR-10,CIFAR1-100,Tiny ImageNet}
    \label{fig:balance_theta_sensitivity}
\end{figure}

\begin{figure}[t]
 \centering
     \includegraphics[keepaspectratio, width=\linewidth]{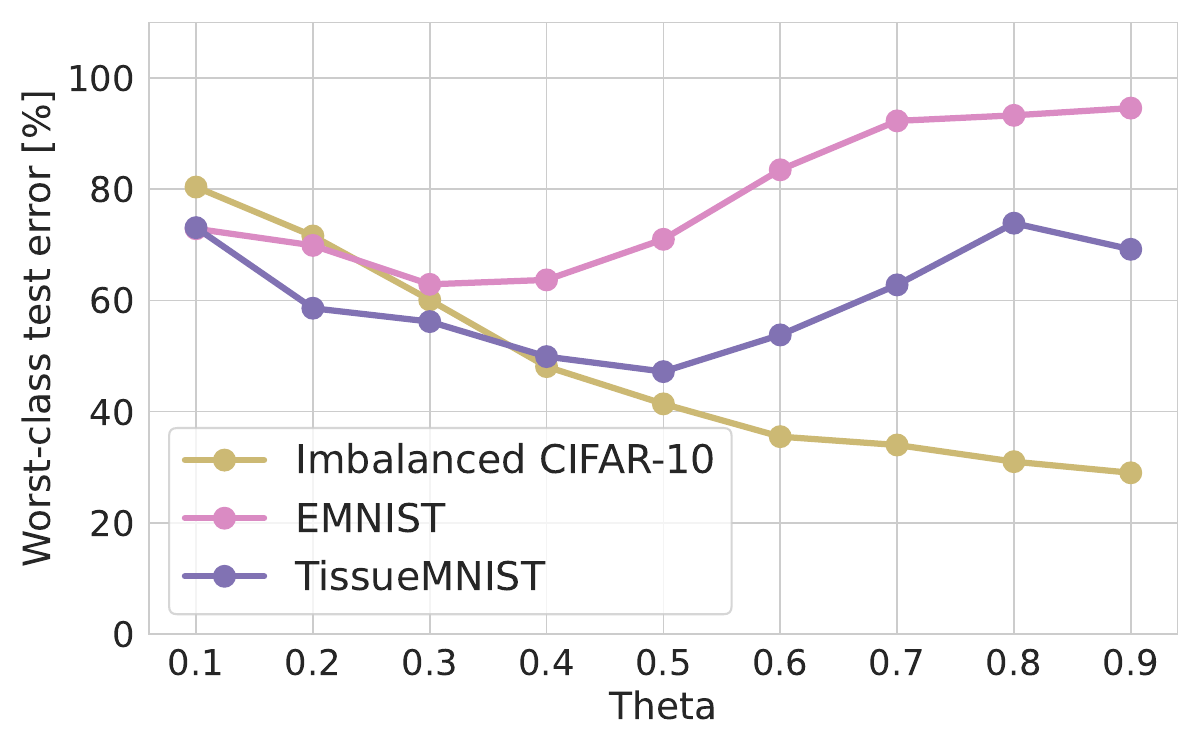}
    \caption{Imbalanced: Imb.CIFAR-10($\rho=10$),EMNIST,TissueMNIST}
    \label{fig:imbalance_theta_sensitivity}
\end{figure}

\begin{table*}
    \caption{Effects of $\theta$ on worst-class and average test errors [\%] on balanced CIFAR-10, CIFAR-100, Tiny ImageNet.} 
    \centering
    \begin{tabular}{lcccccccc}
    \hline
    Balanced dataset & \multicolumn{2}{c}{CIFAR10} &  & \multicolumn{2}{c}{CIFAR100} &  & \multicolumn{2}{c}{Tiny ImageNet} \\ \hline
    Metrics    & Worst         & Avg.        &  & Worst         & Avg.         &  & Worst          & Avg.          \\ \hline \hline
    $\theta = 0.1$   & 95.7          & 67.1        &  & 81.0          & 67.3         &  & 84.0           & 62.4          \\
    $\theta = 0.2$   & 74.9          & 52.5        &  & 75.0          & 58.7         &  & 76.0           & 52.5          \\
    $\theta = 0.3$   & 61.7          & 43.4        &  & 68.0          & 50.4         &  & 68.0           & 45.5          \\
    $\theta = 0.4$   & 54.5          & 35.7        &  & \textbf{62.0}          & \textbf{44.5}         &  & \textbf{62.0}           & 39.8          \\
    $\theta = 0.5$   & 42.2          & 29.9        &  & 89.0          & 49.1         &  & 68.0           & 36.1          \\
    $\theta = 0.6$   & 33.3          & 22.2        &  & 87.0          & 47.4         &  & 70.0           & 34.5          \\
    $\theta = 0.7$   & 28.4          & 17.6        &  & 86.0          & 47.0         &  & 74.0           & \textbf{34.1}          \\
    $\theta = 0.8$   & 22.3          & 12.9        &  & 87.0          & 46.3         &  & 76.0           & 34.6          \\
    $\theta = 0.9$   & \textbf{18.9}          & \textbf{9.6}         &  & 83.0          & 45.4         &  & 80.0             & 37.2            \\ \hline
    \end{tabular}
    \label{tab:theta_balance_sensitivity}
\end{table*}

\begin{table*}
    \caption{Effects of $\theta$ on worst-class and average test errors [\%] on imbalanced CIFAR-10, EMNIST, TissueMNIST.} 
    \centering
    \begin{tabular}{lcccccccc}
    \hline 
    Imb. dataset & \multicolumn{2}{c}{CIFAR10 ($\rho = 10$)} &  & \multicolumn{2}{c}{EMNIST} &  & \multicolumn{2}{c}{TissueMNIST} \\ \hline 
    Metrics      & Worst                & Avg.               &  & Worst        & Avg.        &  & Worst           & Avg.          \\ \hline \hline 
    $\theta = 0.1$     & 80.4                 & 52.8               &  & 72.9         & 18.7        &  & 73.1            & 39.9          \\
    $\theta = 0.2$     & 71.6                 & 47.6               &  & 69.9         & 17.6        &  & 58.6            & 44.1          \\
    $\theta = 0.3$     & 60.1                 & 44.5               &  & \textbf{62.9}         & 17.8        &  & 56.2            & 41.8          \\
    $\theta = 0.4$     & 48.1                 & 35.3               &  & 63.7         & 17.5        &  & 49.9            & 38.9          \\
    $\theta = 0.5$     & 41.4                 & 30.1               &  & 71.0         & 14.6        &  & \textbf{47.2}            & 36.4          \\
    $\theta = 0.6$     & 35.5                 & 24.6               &  & 83.5         & 13.7        &  & 53.8            & 32.8          \\
    $\theta = 0.7$     & 34.0                 & 19.5               &  & 92.3         & 13.6        &  & 62.8            & 36.3          \\
    $\theta = 0.8$     & 31.0                 & 16.4               &  & 93.3         & \textbf{12.6}        &  & 73.9            & \textbf{30.8}          \\
    $\theta = 0.9$     & \textbf{29.0}                 & \textbf{14.9}               &  & 94.6       & 14.9         &  & 69.2            & 42.8          \\ \hline
    \end{tabular}
    \label{tab:theta_imbalance_sensitivity}
\end{table*}

\section{Hyperparameter of Boosting}
We summarize the values of $\gamma$ of Boosting 
in Table~\ref{tab:oco_sets_of_theta}.

\begin{table*}[t]
    \begin{center}
      \caption{Candidates of $\gamma$ of Boosting. }
      \begin{tabular}{l|c}
        \hline
        datasets              & $\gamma$          \\ \hline \hline
        artificial balanced   & 0.0995            \\
        artificial imbalanced & 0.0995            \\
        balanced CIFAR-10     & \{0.1995,0.2995\} \\
        balanced CIFAR-100    & \{0.1995,0.2995\} \\
        balanced TinyImage    & \{0.1995,0.2995\} \\
        imbalanced CIFAR-10   & \{0.1995,0.2995\} \\
        EMNIST                & \{0.1995,0.2995\} \\
        TissueMNIST           & \{0.1995,0.2995\} \\ \hline
      \end{tabular}
          \label{tab:oco_sets_of_theta}
    \end{center}
\end{table*}

\section{Other visualization results}
\subsection{Decision boundaries of all baselines for artificial balanced dataset}
The decision boundary of Naive is shown in Figure~\ref{fig:toy_balance_area}(a).
Although Naive achieved better than other baselines, it seems overfitted compared with ours.
\begin{figure*}[t]
 \centering
  \begin{minipage}{0.23\linewidth}
    \includegraphics[keepaspectratio, width=\linewidth]{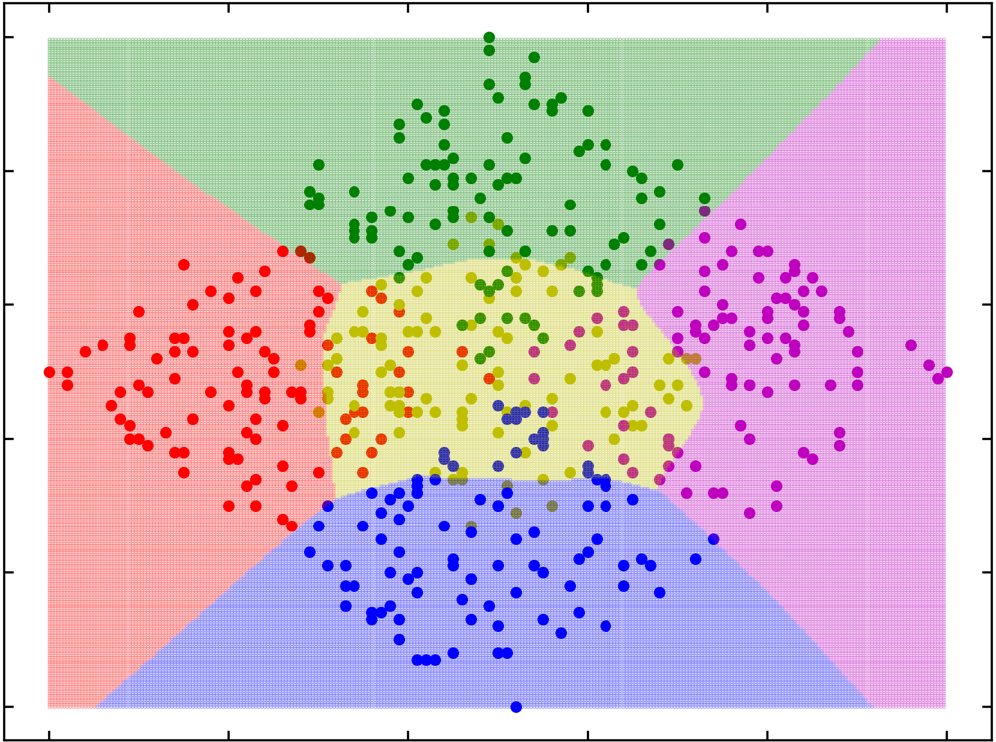}
    \subcaption{Naive}
 \end{minipage}
   \begin{minipage}{0.23\linewidth}
    \includegraphics[keepaspectratio, width=\linewidth]{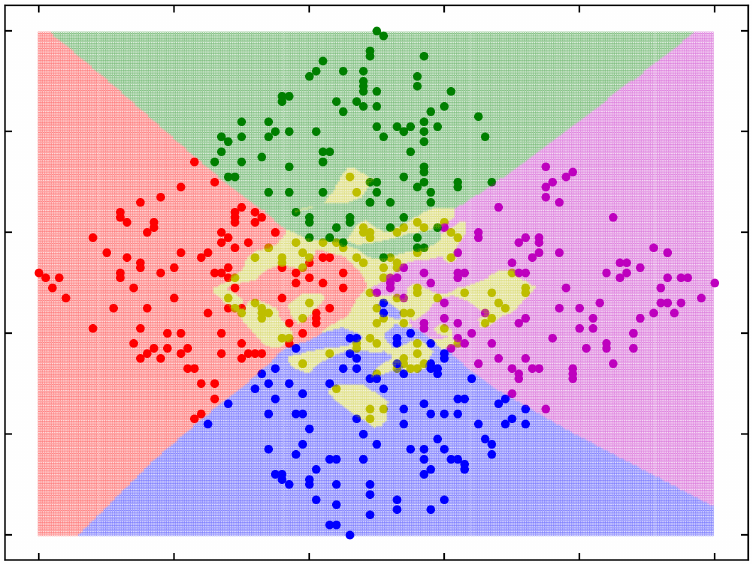}
    \subcaption{Focal}
 \end{minipage}
    \begin{minipage}{0.23\linewidth}
    \includegraphics[keepaspectratio, width=\linewidth]{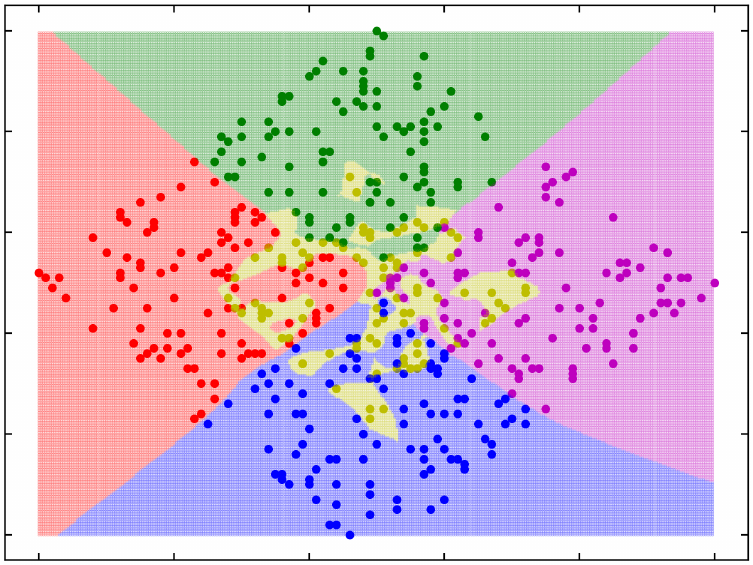}
    \subcaption{IB}
 \end{minipage}
  \begin{minipage}{0.23\linewidth}
    \includegraphics[keepaspectratio, width=\linewidth]{figure/toy_fig/Balanced/area_ours_square.pdf}
    \subcaption{Ours}
 \end{minipage}
 \caption{
 Visualization of the decision boundaries on the balanced artificial dataset.}
 \label{fig:toy_balance_area}
\end{figure*}

\subsection{Decision boundaries artificial imbalanced datasets}
\begin{figure*}[t]
 \centering
    \begin{subfigure}{\linewidth}
    \centering
    \includegraphics[keepaspectratio, width=0.7\linewidth]{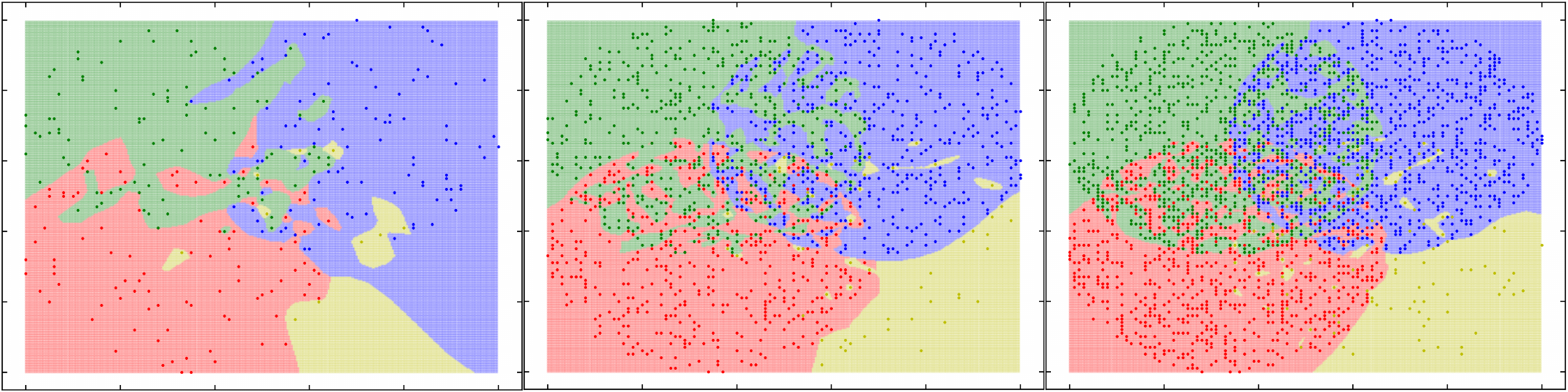}
    \subcaption{CE with $\min_{k \in [K]}{n_k}=10,50,100$} 
    
    \includegraphics[keepaspectratio, width=0.7\linewidth]{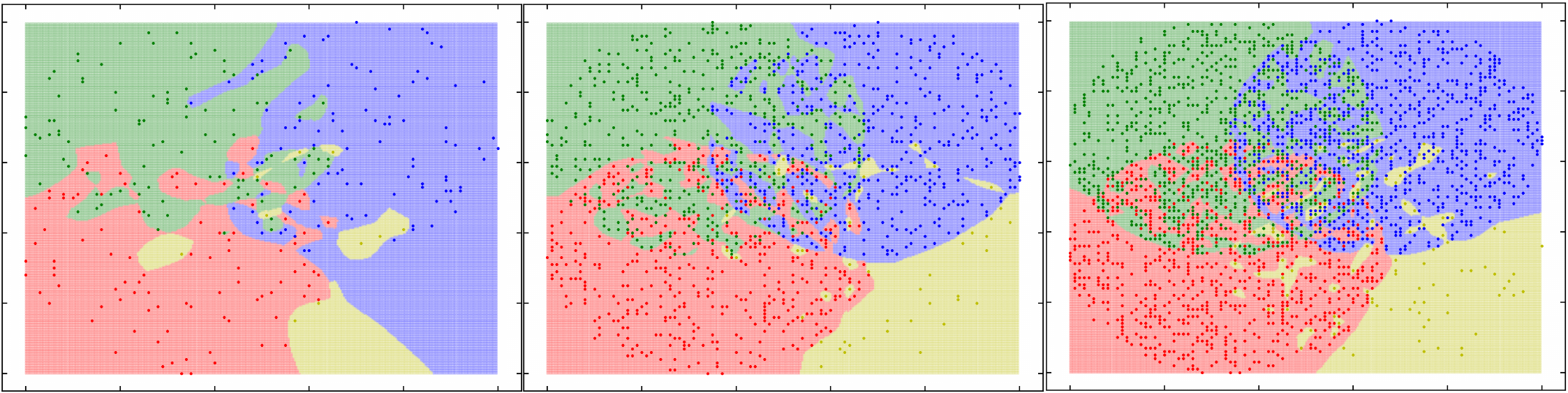}
    \subcaption{wCE with $\min_{k \in [K]}{n_k}=10,50,100$} 

    \includegraphics[keepaspectratio, width=0.7\linewidth]{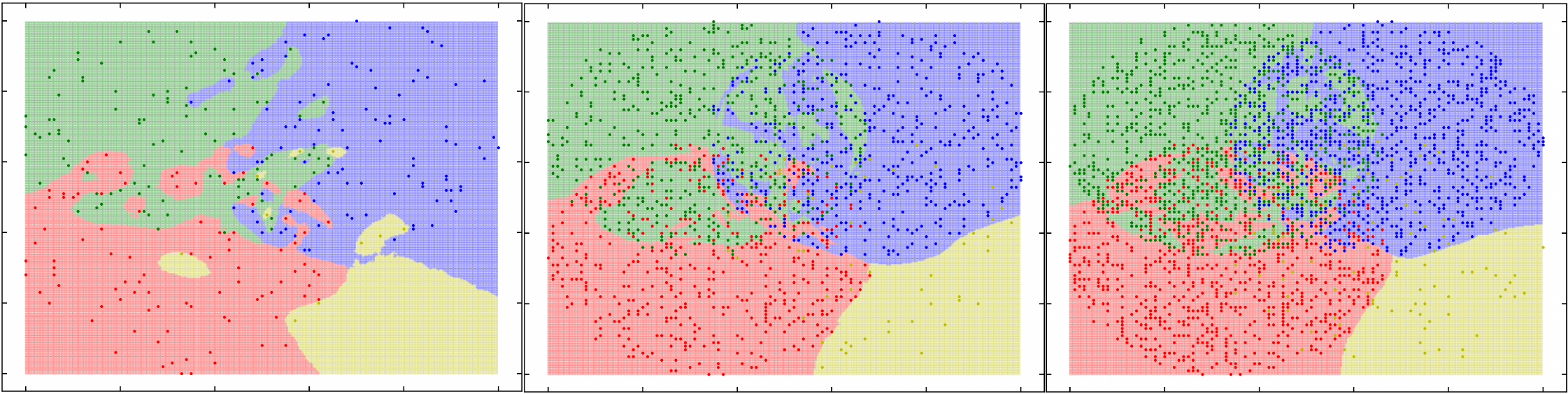}
    \subcaption{Boosting with $\min_{k \in [K]}{n_k}=10,50,100$} 

    \includegraphics[keepaspectratio, width=0.7\linewidth]{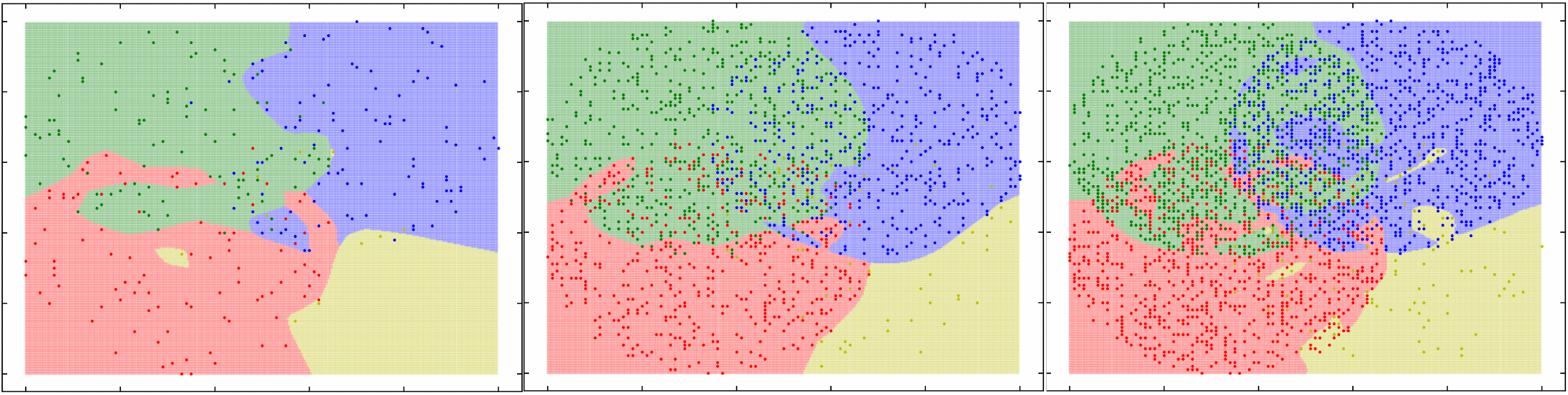}
    \subcaption{0.5DNN with $\min_{k \in [K]}{n_k}=10,50,100$} 

    \end{subfigure}
\caption{Visualization of the decision boundaries on the imbalanced artificial dataset.}
\label{fig:all_class_area_imbalance1}
\end{figure*}

\begin{figure*}[t]
 \centering
    \begin{subfigure}{\linewidth}
    \addtocounter{subfigure}{4}
    \centering
    \includegraphics[keepaspectratio, width=0.7\linewidth]{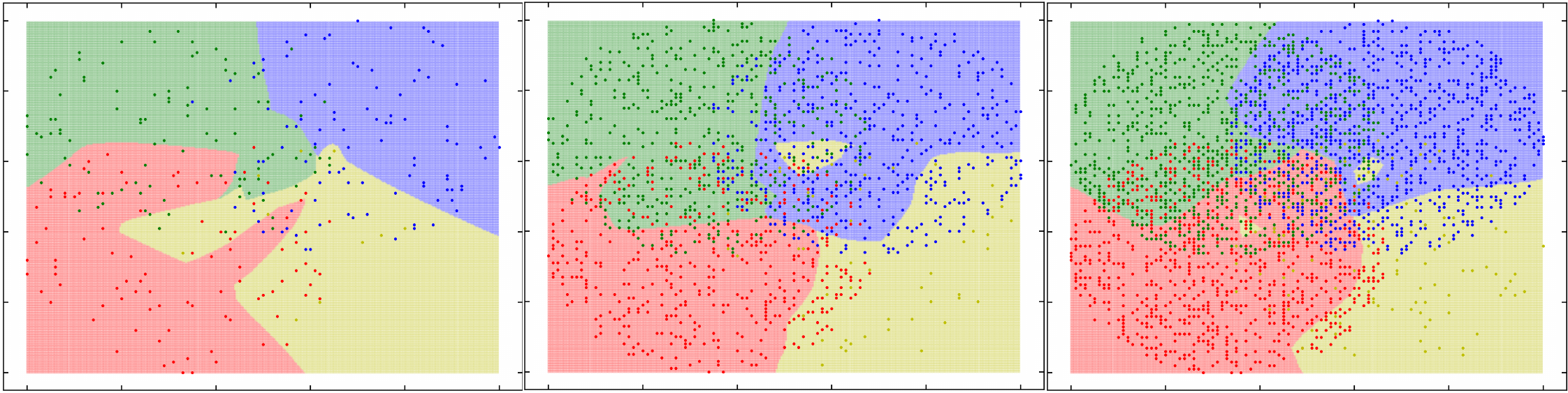}
    \subcaption{Naive with $\min_{k \in [K]}{n_k}=10,50,100$}
    
    \includegraphics[keepaspectratio, width=0.7\linewidth]{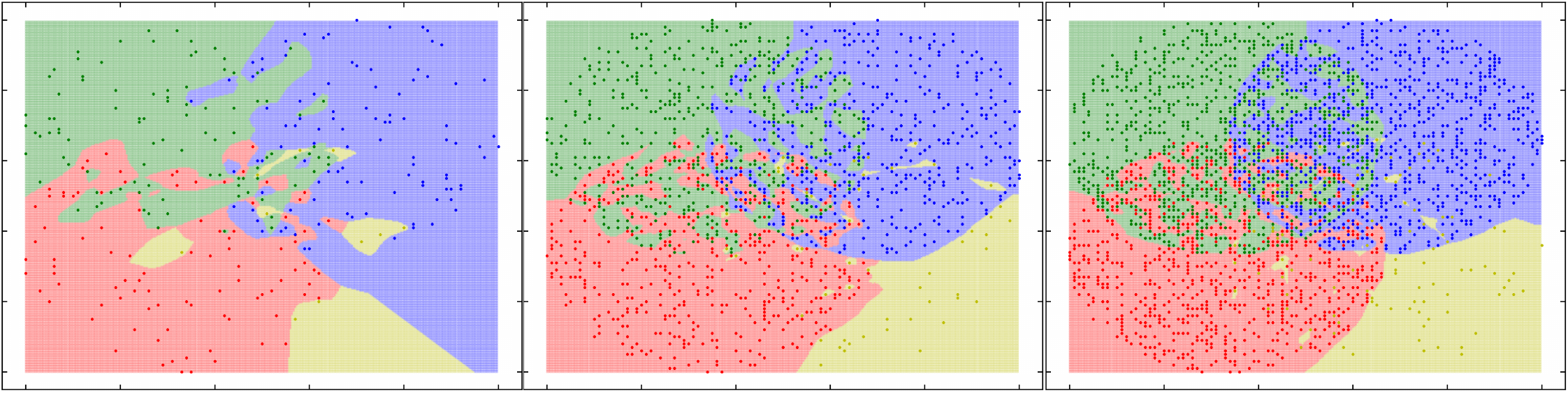}
    \subcaption{Focal with $\min_{k \in [K]}{n_k}=10,50,100$} 
    \includegraphics[keepaspectratio, width=0.7\linewidth]{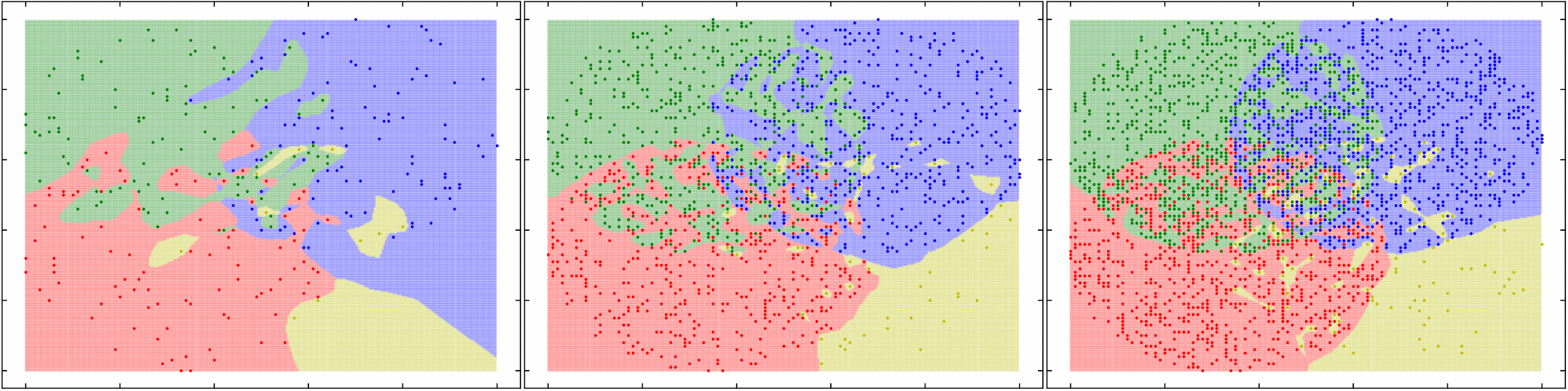}
    \subcaption{CB with $\min_{k \in [K]}{n_k}=10,50,100$} 

    \includegraphics[keepaspectratio, width=0.7\linewidth]{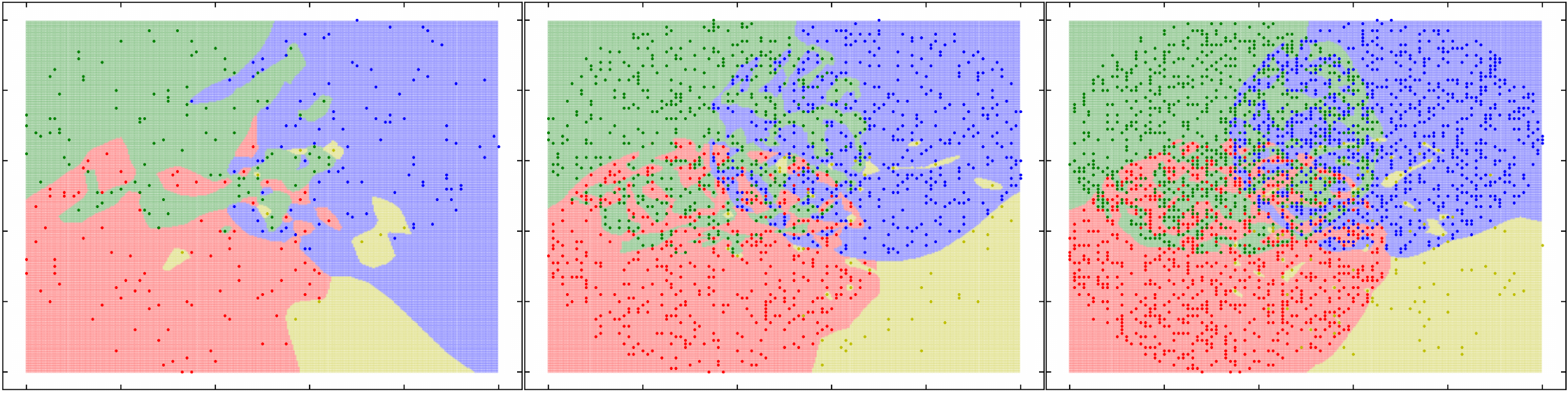}
    \subcaption{IB with $\min_{k \in [K]}{n_k}=10,50,100$} 
    
    \includegraphics[keepaspectratio, width=0.7\linewidth]{figure/toy_fig/Imbalanced/area_ours.pdf}
    \subcaption{Ours with $\min_{k \in [K]}{n_k}=10, 50, 100$}
    \end{subfigure}
    
\caption{Visualization of the decision boundaries on the imbalanced artificial dataset.}
\label{fig:all_class_area_imbalance2}
\end{figure*}

The decision boundary of all baselines is shown in Figure~\ref{fig:all_class_area_imbalance1} and \ref{fig:all_class_area_imbalance2}.
Although Naive achieved better than other baselines, it seems overfitted compared with ours.

\end{appendices}


\clearpage
\clearpage

\end{document}